\documentclass[journal]{IEEEtran}

\usepackage{amsmath,amsfonts}
\usepackage{algorithm}
\usepackage[noend]{algpseudocode}
\usepackage{array}
\usepackage[caption=false,font=normalsize,labelfont=sf,textfont=sf]{subfig}
\usepackage{textcomp}
\usepackage{stfloats}
\usepackage{url}
\usepackage{verbatim}
\usepackage{graphicx}
\usepackage{cite}
\usepackage{multirow, tabularx}
\usepackage{refcount}
\usepackage{makecell}

\DeclareRobustCommand{\svdots}{% s for `scaling'
  \vbox{%
    \baselineskip=0.23333\normalbaselineskip
    \lineskiplimit=0pt
    \hbox{.}\hbox{.}\hbox{.}%
    \kern-0.2\baselineskip
  }%
}

\newcolumntype{P}[1]{>{\centering\arraybackslash}p{#1}}
\newcommand{\LineComment}[1]{\hfill $\triangleright$ \textit{#1}}

\def\ganname{ctdGAN}
\def\numdatasets{14}
\def\numneurons{256}
\def\numseeds{3}

\def\critic{\mathcal{C}}
\def\generator{\mathcal{G}}
\def\clfqy{\mathcal{Q}^y}
\def\clfqu{\mathcal{Q}^u}

\def\lat{\mathbf{z}}
\def\realx{\mathbf{x}}
\def\tranx{\mathbf{x}'}
\def\genx{\mathbf{\hat{x}}}
\def\genu{\mathbf{\hat{u}}}
\def\geny{\mathbf{\hat{y}}}

\def\clfy{\mathbf{y}_Q}
\def\clfu{\mathbf{u}_Q}
\def\mask{\mathbf{m}}

\def\entropy{\mathcal{H}}

\def\relu{\mathsf{ReLU}}

\def\condd{\mathsf{conditions}}
\def\randd{\mathsf{random}}

\def\loss{\mathcal{L}}
\def\prob{\mathbb{P}}
\def\expval{\mathbb{E}}

\begin{document}

\title{A Conditional GAN for Tabular Data Generation with Probabilistic Sampling of Latent Subspaces}

\author{Leonidas Akritidis, Panayiotis Bozanis
\thanks{L. Akritidis is with the Department of Information and Electronic Engineering, 
International Hellenic University, Thessaloniki, Greece.
P. Bozanis is with the School of Science and Technology, International Hellenic University, 
Thessaloniki, Greece.}}

\markboth{IEEE Transactions on Knowledge and Data Engineering,~Vol.~0, No.~0, August~2026}%
{How to Use the IEEEtran \LaTeX \ Templates}

\IEEEpubid{0000--0000/00\$00.00~\copyright~2021 IEEE}

\maketitle

\begin{abstract}
The tabular form constitutes the standard way of representing data in relational database systems
and spreadsheets. But, similarly to other forms, tabular data suffers from class imbalance, a
problem that causes serious performance degradation in a wide variety of machine learning tasks.
One of the most effective solutions dictates the usage of Generative Adversarial Networks (GANs) in
order to synthesize artificial data instances for the under-represented classes. Despite their good
performance, most of the proposed GAN models do not take into account the vector subspaces of the
input samples in the real data space, leading to data generation in arbitrary locations. In
addition, the class labels are handled in the same manner as the other categorical variables, so
conditional sampling by class is rendered less effective. To overcome these problems, this study
presents \ganname{}, a conditional GAN for alleviating class imbalance in tabular datasets.
Initially, \ganname{} executes a space partitioning step to assign cluster labels to the input
samples. Subsequently, it utilizes these labels to synthesize samples via a novel probabilistic
sampling strategy and a new loss function that penalizes both cluster and class mis-predictions. In
this way, \ganname{} generates samples in subspaces that resemble those of the original data
distribution. We also introduce several other improvements, including a simple, yet effective
cluster-wise scaling technique that captures multiple feature modes without affecting data
dimensionality. The evaluation of \ganname{} with \numdatasets{} imbalanced datasets demonstrated
its strong ability in generating high fidelity samples and improving classification accuracy.
\end{abstract}


\section{Introduction}
\label{sec:intro}

Data comes in many forms. The tabular one is among the most important, due to its presence in
numerous problems across multiple disciplines, such as Engineering, Physics, Biology, Finance, etc.
Nowadays, massive amounts of tabular data are generated, either manually or automatically, by
sensors, instruments, software, bots, and, of course, humans.

The availability of such volumes of data triggered the introduction of countless applications with
the aim of learning the underlying data distribution and making predictions with it. Indicative
examples include fault predictors \cite{eswa2021c,is2024}, power consumption estimators
\cite{ictai2017}, stock market price predictors \cite{ent2020}, intrusion detectors \cite{jbd2021},
anomaly detectors \cite{tkde2021}, and so on.

One of the most significant problems in these applications is class imbalance; that is, the uneven
distribution of the input samples in the involved classes. The consequences of class imbalance in
the performance of a predictor are critical \cite{tkde2015}. The model becomes biased towards the
majority class and cannot learn the other classes effectively \cite{ictai2023}. For this reason, it
is imperative that the problem is confronted before training, so that an effective estimator can be
obtained.

During the past years, numerous approaches for alleviating class imbalance have been proposed
\cite{tkde2009}. Among them, the oversampling techniques synthesize artificial samples with the aim
of augmenting an imbalanced dataset \cite{tkde2015}. Traditional methods like SMOTE are still being
utilized extensively, mainly due to their inherent simplicity and effectiveness \cite{jair2002}.

The rapid advances in deep learning research led to the introduction of state-of-the-art generative
models such as Variational Autoencoders (VAEs) \cite{arxiv2013}, Generative Adversarial Networks
(GANs) \cite{anips2014}, and Diffusion models \cite{anips2021}. In many cases, the excellent
performance of GANs rendered them attractive candidates for oversampling tasks. A GAN consists of
two networks trained jointly in an adversarial fashion: the Generator and the Discriminator. The
former takes as input random samples from a latent distribution and transforms them into meaningful
data instances. On the other hand, the Discriminator is trained to distinguish between real and
synthetic data produced by the Generator.

The Conditional GAN (CGAN) extended this functionality by enabling the generation of samples
belonging to a particular class \cite{arxiv2014}. Later, the Wasserstein GAN (WGAN) replaced the
Discriminator by a Critic, a network that estimates how far from reality the synthetic samples,
created by the Generator, are \cite{anips2017,icml2017}. Nevertheless, WGAN is not conditional.

\IEEEpubidadjcol

Regarding tabular data, TableGAN takes privacy into account by synthesizing data without leaking
sensitive information \cite{vldb2018}. ctGAN employs a flexible Generator that produces both
numeric and categorical variables, conditioned by any discrete column \cite{anips2019}. CTAB-GAN
encodes mixtures of continuous and discrete values and introduces a loss function that embodies
multiple types of errors \cite{acml2021}. CG-TGAN replaces the dense (or convolutional) layers of a
GAN with graph neural networks to generate more realistic data \cite{aaai2025}.

In a new line of research, TabDDPM adapts the denoising diffusion probabilistic model to the domain
of tabular data synthesis \cite{icml2023}. In general, the training of DDPMs is more stable, but
also computationally more expensive than that of GANs. The inference phase requires multiple 
forward passes, rendering data generation slower. In addition, DDPMs usually require large amounts
of data to be effectively trained; this is not always the case in applications involving tabular
data.

All these models ignore the issue of data locality, namely, the vector subspace where an input
sample originally belongs. Therefore, they cannot directly affect the location (in the original
vector space) of their generated samples, increasing the risk of distorting the respective class
distributions. Additionally, they treat the class labels in the same manner as the other discrete
variables, failing to quantify mis-classifications in the underlying loss functions.

To address these problems, we present a new model for tabular data synthesis, named \ganname{}. The
training of \ganname{} begins with a space partitioning step that assigns cluster labels to the
input samples. During the training loop, the Generator samples from a latent mixture of normally
distributed continuous variables, random discrete variables and random cluster and class labels.
The creation of mis-classified and mis-clustered samples is subsequently penalized by a novel loss
function and two auxiliary classifiers. The random sampling from different clusters combined with
the heavier penalization forces \ganname{} to generate samples with reasonable diversity, reducing
the risk of mode collapse.

During data synthesis, \ganname{} applies a new probabilistic sampling technique that, apart from
the requested class, also selects the most appropriate clusters to create samples into. In this
context, \ganname{} does not just create a sample from a particular class, but it also identifies
the most promising subspaces (i.e. clusters) where that sample should be generated. In this way,
the model generates data that better fits the original data distribution.

In brief, the contributions of this work include:

\begin{itemize}
\item{\ganname{}: a conditional GAN that combines the classes and the subspaces of the input data
in the original space to effectively learn the underlying distribution. The model establishes a
novel latent space from where the Generator samples latent mixtures of continuous values, discrete
variables, and random cluster and class labels.}

\item{Generator Loss function: We propose a new loss function that: i) incorporates the synthetic 
data quality, and ii) penalizes the generation of unrealistic classes and clusters that are
different from the respective latent ones.}

\item{Probabilistic sampling: The model maintains a conditional probability matrix whose elements
represent the probability that a sample belongs to a cluster, given its class. When \ganname{} is
requested to synthesize samples from a specific class, this matrix indicates suitable clusters to
place these samples into.}

\item{Cluster-wise scaling: We adopt a simple method for scaling the continuous features in a way
that captures different modes without increasing dimensionality, in contrast to other cluster-based
normalizers such as VGM.}

\item{We extensively evaluate the performance of \ganname{} on a broad collection of \numdatasets{}
datasets.}
\end{itemize}

The rest of the paper is organized as follows: Section \ref{sec:rwork} cites the most significant
advances in the area of GAN-based tabular data synthesis. Section \ref{sec:gan} describes the
architecture and the basic components of \ganname{}. The model performance is evaluated in Section
\ref{sec:exp} and discussed in Section \ref{sec:disc}. Finally, Section \ref{sec:conc} states the
conclusions of this study and outlines the most significant insights for future work.

\section{Related Work}
\label{sec:rwork}

The Generative Adversarial Network (GAN) was introduced by Goodfellow et al. \cite{anips2014}. It 
adopts an adversarial learning strategy, where two networks compete with each other. At each
iteration, the Generator synthesizes samples whose quality is subsequently judged by a classifier,
called the Discriminator. It is a zero-sum game, where both networks gradually improve their
performance. After training, the Generator can synthesize data instances by sampling a normally
distributed latent space.

The Conditional GAN was introduced with the aim of creating data instances belonging to a specific
class \cite{arxiv2014}. By concatenating the input feature vectors with their one-hot encoded class
labels, the Generator learns how to create samples belonging to a particular class. InfoGAN
extended this idea by maximizing the mutual information between the latent and the real data
\cite{anips2016}.

Subsequently, Wasserstein GAN (WGAN) improved the training stability of vanilla GANs by replacing
the Discriminator with a Critic \cite{icml2017}. The Critic does not discriminate between real and
synthetic data, but outputs a value that represents how realistic the generated samples are. WGAN
alleviates the problem of mode collapse that occurs when a GAN repeatedly generates samples of
limited diversity.

In the domain of tabular data, ctGAN conditionally synthesizes data having both continuous and
discrete columns \cite{anips2019}. The Generator's output layer has a dual activation function:
hyperbolic tangent for the continuous values, and Gumbel-softmax for the discrete ones. To capture
multiple modes in the distributions of the continuous columns, ctGAN employs a Variational Gaussian
Mixture Model (VGM) to normalize the data. Unfortunately, VGM outputs high dimensional values that
negatively affect training times. In contrast, \ganname{} uses a simple cluster-based normalization
strategy that captures different distribution modes without affecting dimensionality.

Furthermore, TableGAN was designed to synthesize tabular data without incurring information leakage
\cite{vldb2018}. Based on DCGAN \cite{arxiv2015}, the model handles both categorical and numeric
columns by using convolutional architectures. It also maintains a classifier to predict the 
class of the synthetic samples. In contrast, ehrGAN utilized Recurrent Neural Networks (RNNs) with
the aim of generating electronic health records with sensitive information \cite{icdm2017}.
CTAB-GAN also synthesizes sensitive tabular data by considering columns of mixed data types 
\cite{acml2021}.

The requirement for differential privacy during tabular data generation has attracted numerous
researchers. One way of preventing the leakage of sensitive variables is by combining a 
differential private version of stochastic gradient descent with Wasserstein loss. This strategy
has been adopted, in one form or another, by multiple relevant models, including DPGAN
\cite{arxiv2018c}, GS-WGAN \cite{anips2020}, and others.

Several GANs have been especially designed to improve classification performance in imbalanced
datasets. For example, SB-GAN capitalizes on the idea of training by using important samples only.
The model first discards the outliers and then, it assigns higher weights to the samples that lie
near the decision boundary. In contrast, IDA-GAN employs a VAE to learn the distributions of the
majority and minority classes in the latent space and uses them during training \cite{icpr2021}.

\section{Tabular Data Generation with \ganname}
\label{sec:gan}

\begin{figure*}[!t]
\center
\includegraphics[width=\linewidth]{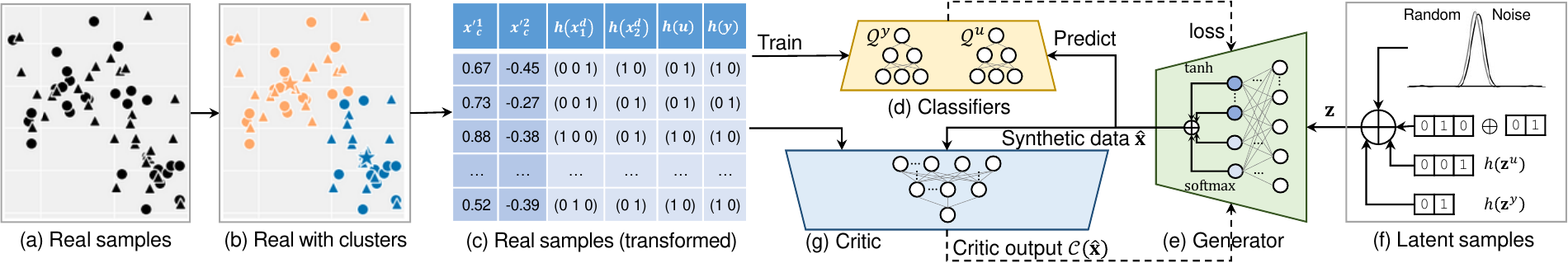}
\caption{
\ganname{} architecture: (a) The input data space: different marker shapes denote different
classes. (b) Space-partitioned samples: different colors denote different subspaces (clusters). (c)
The transformed data instances: columns with dark (light) background represent continuous 
(discrete) variables. (d) The auxiliary classifiers $\clfqy$ and $\clfqu$ that judge the class and cluster labels of the generated samples. (e) The Generator network $\generator$ with its dual activation function: Gumbel 
softmax for the discrete variables and tanh for the continuous. (f) The latent space from where the
$\generator$ takes samples. (g) The Critic $\critic$ receives both real and synthetic samples.}
\label{fig:ctd-gan}
\end{figure*}

This section presents the proposed \ganname{} model for tabular data synthesis. It describes the
initial clustering step, the subsequent data transformations, the architecture and training of the
model, and, finally, the probabilistic sampling algorithm. Figure~\ref{fig:ctd-gan} depicts the
involved components and procedures.

\subsection{Clustering and Real Space Setup}
\label{ssec:data-prep}

Let $\mathbf{X}$ be a tabular dataset with $m$ rows and $n$ columns. We refer to the columns with
discrete and continuous values as $X^d_1,X^d_2,\dots,X^d_{n_d}$ and $X^c_1,X^c_2,\dots,X^c_{n_c}$
respectively, where $n_d$ and $n_c$ denote the number of discrete and continuous columns. Each row
$\mathbf{x}_i\in\mathbf{X}$ is associated with a target variable $y_i\in Y$ that represents its
class label, where $Y$ is the set of all classes. These elements suffice to define $\realx_i$:
\begin{equation}
\realx_i=\big(x^c_{i,1},\dots,x^c_{i,n_c},x^d_{i,1},\dots,x^d_{i,n_d},y_i\big).
\label{eq:vy}
\end{equation}

Most competitive models (e.g., ctGAN, TableGAN, etc.) do not explicitly take into consideration the
target variables $y_i$. Instead, they embed them into $\realx_i$, treating them as typical
categorical features. In the following subsections we show that the separation of $y_i$ from the
rest of the categorical features yields substantial improvements in the generated data quality.

The training of \ganname{} begins with a clustering algorithm that assigns a cluster label $u_i$ to
each row of $\mathbf{X}$. After that, $u_i$ is embedded into the representation of Eq. \ref{eq:vy},
leading to:
\begin{equation}
\mathbf{x}_i=\big(x^c_{i,1},\dots,x^c_{i,n_c},x^d_{i,1},\dots,x^d_{i,n_d},u_i,y_i\big).
\label{eq:vcy}
\end{equation}

Eq. \ref{eq:vcy} imposes each input sample $\mathbf{x}_i\in\mathbf{X}$ to be represented by: i) its
feature vector, comprising both discrete and continuous components, ii) the label of its
respective cluster $u_i$, and iii) its class label $y_i$.

Regarding the clustering algorithm, \ganname{} employs hierarchical Agglomerative clustering with
complete linkage to identify groups of similar samples in mixed-type datasets. Since tabular data
comprises variables of different scales, pairwise dissimilarities are computed using the Gower
distance metric \cite{bio1966}. Gower distances are designed to accommodate heterogeneous data
types, including continuous, categorical, and binary variables. The resulting clusters reflect
patterns of similarity across the full set of variables while preserving the multidimensional
nature of the mixed-type data.

The ideal number of clusters to be constructed, $k$, is determined by a simple heuristic that
minimizes the value of scaled inertia:
\begin{equation}
SI=\frac{\sum^k_{u=1}I_u(k)}{I(k=1)}+\alpha k,
\label{eq:si}
\end{equation}
\noindent where $I_u(k)$ is the inertia of a cluster $u$, calculated by summing the Gower distances
of each sample from its cluster medoid. $I(k=1)$ is the inertia without clustering (i.e., $k=1$),
whereas $\alpha\in[0.01, 0.1]$ is a parameter that penalizes the number of clusters $k$. Here we
set it equal to $\alpha=7\cdot 10^{-2}$.

\subsection{Column Transformations}
\label{ssec:trans}

Data scaling is of crucial importance when training machine learning models. Without it, a feature
with significantly larger values might dominate the objective function, preventing the model from
learning from other features.

In \cite{anips2019} the authors highlighted the existence of multiple modes in almost half of the
continuous columns of their test datasets. They proposed a Variational Gaussian Mixture Model (VGM)
that is capable of capturing multiple modes in the distribution of the continuous variables.
However, VGM outputs not only the normalized continuous values, but also the labels of their
respective mixture components. Subsequently, the discrete component labels are one-hot encoded,
leading to very high-dimensional representations of the transformed continuous values.

To confront this problem, we introduce a simple technique that exploits the cluster labels of the
previous subsection. Specifically, we perform min-max normalization, but only \textit{relatively to
the samples that belong to the same cluster $u_i$}. In this context, to transform a continuous
variable $x^c_{i,j}$, we first compute the minimum and maximum values from the samples belonging to
the same cluster as $\realx_i$. Then, we scale in the range $[-1,1]$, so that the normalized values
correspond to the $\tanh$ activation function of the Generator:
\begin{equation}
x'_{i,j} = 2\frac{ x^c_{i,j} - \min{V_{i,j}} } { \max{V_{i,j}} - \min{V_{i,j}} } - 1,~~
V_{i,j}\subseteq\{X^c_j|u_i\}
\label{eq:mms}
\end{equation}
\noindent where $i$, $j$ represent the row and the column of $x^c_{i,j}$, respectively. The subset
$V_{i,j}\subseteq X^c_j$ contains all the values of the $j$-th continuous column $X^c_j$ that also
belong to cluster $u_i$.

Similarly to all cluster-based normalizers (like VGM), Eq.~\ref{eq:mms} may assign similar scaled
values to variables that are actually different. For example, the maximum value of a feature among
the samples of the same cluster will always be transformed to $1$ (and the minimum to $-1$); this 
applies to all clusters. Consequently, it is crucial that the Generator quickly learns to output 
correct cluster labels, so that the transformation can be inverted correctly. This requirement is
satisfied by appropriately structuring the loss function of the Generator.

The proposed method addresses the aforementioned challenges, since it transforms a continuous 
column in a way that: i) does not append redundant columns to the dataset (as VGM does), and ii)
captures different modes in the underlying data distribution, due to its cluster-aware nature.

Regarding the discrete columns $X^d$, \ganname{} applies a typical one-hot transformation $h(X^d)$.
One-hot encoding is also applied to the class and cluster labels.

After these transformations take place, the vector representation $\tranx_i$ of an input sample
$\realx_i$ is defined by the concatenation ($\oplus$) of the transformed continuous and discrete
variables, with its one-hot cluster and class labels:
\begin{equation}
\begin{aligned}
\tranx_i = &\big(x'_{i,1},\dots,x'_{i,n_c}\big) \oplus          && \textit{continuous columns}   \\
           &\big(h(x^d_{i,1}),\dots,h(x^d_{i,n_d})\big) \oplus  && \textit{discrete columns}     \\
           & h(u_i) \oplus h(y_i).                              && \textit{cluster/class labels}
\end{aligned}
\label{eq:evcy}
\end{equation}
Due to the length of the one-hot discrete columns $|h(\cdot)|$, the dimensionality of the
transformed samples now increases to $|\tranx_i|=n_c+\sum_{i=1}^{n_d}|h(X^d_i)|+k+|h(y_i)|$.

\subsection{Latent Space}
\label{ssec:lat-space}

Conditional GANs (CGANs) extend the functionality of vanilla GANs by supporting the generation of
data samples belonging to a particular class. This is achieved by first sampling from a prior that
comprises normally distributed variables $\lat^c~\sim\mathcal{N}(0,\sigma)$. In the sequel,
$\lat^c$ is concatenated with a random latent target variable $\lat^y$ and it is fed to the
Generator. The model is trained to optimize a loss function that penalizes the generated samples
that have $y\neq \lat^y$.

In \ganname{} the latent space is formed as follows: Initially, the Generator randomly samples
from an $e$-dimensional prior $\lat_{i}^{c}~\sim\mathcal{N}(\mu,\sigma),~\mu=0,~\sigma=1,~e=128$ to
obtain a normally distributed latent vector $\lat^c$. 

Next, a conditional vector $\lat^d$ is formed: For each discrete column $\{X_i^d\}_{i=1}^{n_d}$, we
create an $|X_i^d|$-dimensional mask vector $\mask_i$ and we initialize it to 0. Then, we randomly
select a discrete column $i^*$ (including the class and cluster columns) and a value $\lat^d_{i^*}
\in X_{i^*}^d$. Subsequently, the $i^*$-th mask vector $\mask_{i^*}$ is set equal to the one-hot
representation of the selected value, so that $\mask_{i^*}=h(\lat^d_{i^*})$. Now the conditional
vector $\lat^d$ is a concatenation of the computed mask vectors:
\begin{equation}
\lat^d = \mask_1 \oplus \dots \oplus \mask_{n_d} \oplus \mask_u \oplus \mask_y.
\label{eq:condvec}
\end{equation}
\noindent that, in turn, is concatenated with $\lat^c$ to obtain the final representation of the
latent samples $\lat$:
\begin{equation}
\begin{aligned}
\mathbf{z} &= \lat^c \oplus \lat^d \\
           &= \lat^c \oplus \mask_1 \oplus \dots \oplus \mask_{n_d} \oplus \mask_u \oplus \mask_y.
\end{aligned}
\label{eq:latsample}
\end{equation}
The vector $\lat^d$ explicitly presents the condition $X_{i^*}^d=\lat^d_{i^*}$ to the Generator,
strengthening its ability to conditionally generate samples by any discrete column. This is of
crucial importance in downstream classification tasks, where the model is requested to generate
samples from specific classes.

\begin{figure}[!t]
\center
\includegraphics[width=\linewidth]{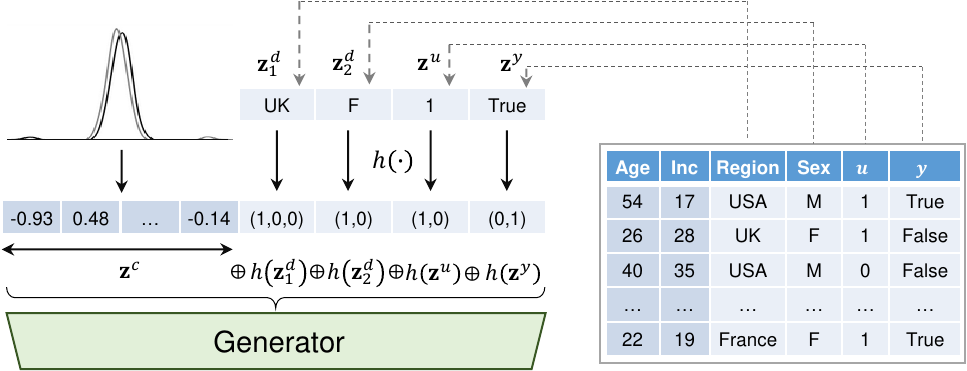}
\caption{Creation of a latent sample $\lat$, composed by: i) an $e$-dimensional latent vector
$\lat^c$ sampled from a normal distribution, ii) $n_d=2$ latent discrete variables $\lat^d$, iii) a
random cluster label $\lat^u$, and iv) a random class label $\lat^y$.}
\label{fig:zsamp}
\end{figure}
Eq.~\ref{eq:latsample} determines the input of the Generator. Its dimensionality is
$|\lat|=e+\sum_{i=1}^{n_d}|h(\lat^d_i)|+k+|h(\lat^y)|$ and defines the size of the Generator's
input layer. These chained concatenations are visualized in Fig.~\ref{fig:ctd-gan} (block (f)) and
the exemplary Fig.~\ref{fig:zsamp}.

\subsection{Architecture}
\label{ssec:arch}

\ganname{} adopts a fairly simple architecture where both the Generator $\generator$ and the Critic
$\critic$ are implemented as fully-connected feed-forward networks. More specifically, the Critic
$\critic$ includes two hidden layers of size \numneurons{}. The output of each layer passes through
Leaky ReLU activation and a Dropout layer with probability 0.2 \cite{jmlr2014}. The input samples 
adopt the form of Eq.~\ref{eq:evcy} and are fed to $\critic$ in packs of $pac=10$ in order to
prevent mode collapse (according to PacGAN \cite{anips2018}). 

The Generator $\generator$ includes two residual blocks and each block contains one fully-connected
layer with \numneurons{} neurons and ReLU activation. We also include one batch normalization layer
at the end of each residual block to reduce the internal covariate shift during training
\cite{icml2015}.

The model receives its input $\lat$ from the latent space according to Eq.~\ref{eq:latsample}.
Regarding the output $\genx=\generator(\lat)$, we adopt the inspiring strategy of ctGAN 
\cite{anips2019}; the continuous values are generated by a hyperbolic tangent function $\tanh$,
whereas the discrete ones are generated by Gumbel softmax \cite{arxiv2016}.

In addition, \ganname{} implements two auxiliary classifiers $\clfqy$ and $\clfqu$ to provide 
stable and differentiable class and cluster-conditioning signals to the Generator during
adversarial training. They are not intended to serve as fully optimized predictive models, but
rather as guidance networks that provide informative supervision to the Generator. If $\clfqy$ and
$\clfqu$ were trained until near-perfect convergence, their posterior distributions would become
highly peaked, causing the corresponding cross-entropy losses to approach zero for most generated
samples. Consequently, the gradients propagated to the Generator through the auxiliary losses would 
become weak and provide little corrective information.

To maintain informative supervision, the classifiers are pre-trained for at most 30 epochs, which
was empirically found to achieve high predictive performance without driving the posterior
distributions to saturation. This strategy preserves sufficiently accurate class and cluster
guidance while maintaining informative gradients throughout adversarial training.

The classifiers are pre-trained (i.e. before the main adversarial loop) with the same training 
examples as those of \ganname{}. $\clfqy$ uses the sample classes as target variables, whereas
$\clfqu$ accepts their respective cluster labels. They adopt identical MLP structures with two
fully-connected hidden layers of \numneurons{} dimensions, followed by a Dropout layer (with
probability 0.2) and Leaky ReLU activation (with slope 0.2). Both networks use the Adam optimizer
to minimize the cross-entropy loss function.

\subsection{Loss Functions}
\label{ssec:loss}

The training of \ganname{} is based on the principles of the WGAN-GP model \cite{anips2017}. In
general, WGANs are hard to train because the technique of weight clipping affects loss computation,
often leading to vanishing or exploding gradients. WGAN-GP addresses this problem by penalizing the
norm of the Critic's output gradient with a loss function of the form:
\begin{equation}
\loss_\critic =
\expval_{\genx\sim\prob_g} [ \critic(\genx) ] -
\expval_{\realx\sim\prob_r} [ \critic(\realx) ] +
\loss_{\mathcal{GP}} ( \genx, \prob_{\genx} ),
\label{eq:c-loss}
\end{equation}
\noindent where $\prob_r$ and $\prob_g$ are the distributions of the real and generated samples,
respectively. Moreover, $\prob_{\genx}$ is the sampling distribution, formed by uniformly taking
samples in straight lines between $\prob_r$ and $\prob_g$. The third term,
$\loss_{\mathcal{GP}}(\genx,\prob_{\genx})$, is the gradient penalty coefficient, defined in
\cite{anips2017}:
\begin{equation}
\label{eq:gradpen}
\loss_{\mathcal{GP}} (\genx, \prob_\genx) = \lambda
\expval_{\genx\sim\mathbb{P}_{\genx}}\big[(\lVert\nabla_{\genx}\critic(\genx)\rVert_2 -1)^2\big].
\end{equation}

Regarding the Generator $\generator$, \ganname{} introduces the following novel loss function:
\begin{equation}
\label{eq:g-loss}
\loss_\generator =
-\expval_{\genx\sim\mathbb{P}_g} [ \critic(\genx) ] +
\sum_{i=1}^{n_d}\entropy(\lat^d_i, \genx^d_i) + \loss_\mathbf{y} + \loss_\mathbf{u},
\end{equation}
where $\entropy$ is the cross-entropy loss. Eq.~\ref{eq:g-loss} includes four terms. The first
two are also used by ctGAN, CTAB-GAN, etc. and penalize respectively: i) the synthetic data quality
(evaluated by $\critic$), and ii) the generation of inconsistent values for the discrete columns
(compared to the latent ones $\lat^d_i$). However, in this way the class and cluster labels are
considered equivalent to any other discrete variable. To cure this weakness, \ganname{} introduces
the other two terms, $\loss_\mathbf{y}$ and $\loss_\mathbf{u}$, defined as follows:
\begin{equation}
\label{eq:gy-loss}
\loss_\mathbf{y} = \loss_\mathbf{y}(\lat^y, \geny, \clfy) = \entropy(\lat^y, \geny) + \lambda_Q \entropy(\clfy,\geny)
\end{equation}
\begin{equation}
\label{eq:gu-loss}
\loss_\mathbf{u} = \loss_\mathbf{u}(\lat^u, \genu, \clfu) = \entropy(\lat^u, \genu) + \lambda_Q \entropy(\clfu, \genu)
\end{equation}
where $\clfy=\clfqy(\genx)$ and $\clfu=\clfqu(\genx)$ respectively represent the class and cluster
labels of the generated samples $\genx$, as they are predicted by the auxiliary classifiers
$\clfqy$ and $\clfqu$. Eq.~\ref{eq:gy-loss} applies a twofold penalization to $\generator$, namely,
the generation of samples with classes $\geny$ that are: i) different than the respective latent
ones $\lat^y$, and/or ii) do not agree with the judgement of $\clfqy$. In the same spirit, Eq.
\ref{eq:gu-loss} penalizes the creation of samples with incorrect and unrealistic cluster labels.

The parameter $\lambda_Q\in [0,1]$ is used to regulate the weight of the classifiers' predictions.
For $\lambda_Q=0$, the predictions of $\clfqy$ and $\clfqu$ are essentially ignored and the loss
function of $\generator$ is reduced to that of ctGAN. Here we initially assign it a very small 
value (e.g. $\lambda_Q=10^{-2}$) and we keep it fixed for a warm-up period of 20 epochs, so that
$\generator$ learns to generate meaningful data instances. After this period, $\lambda_Q$
progressively increases at a rate of $2\%$ per epoch until it reaches a maximum allowed value of
$1$.

\begin{figure}[!t]
\center
\includegraphics[width=.49\linewidth]{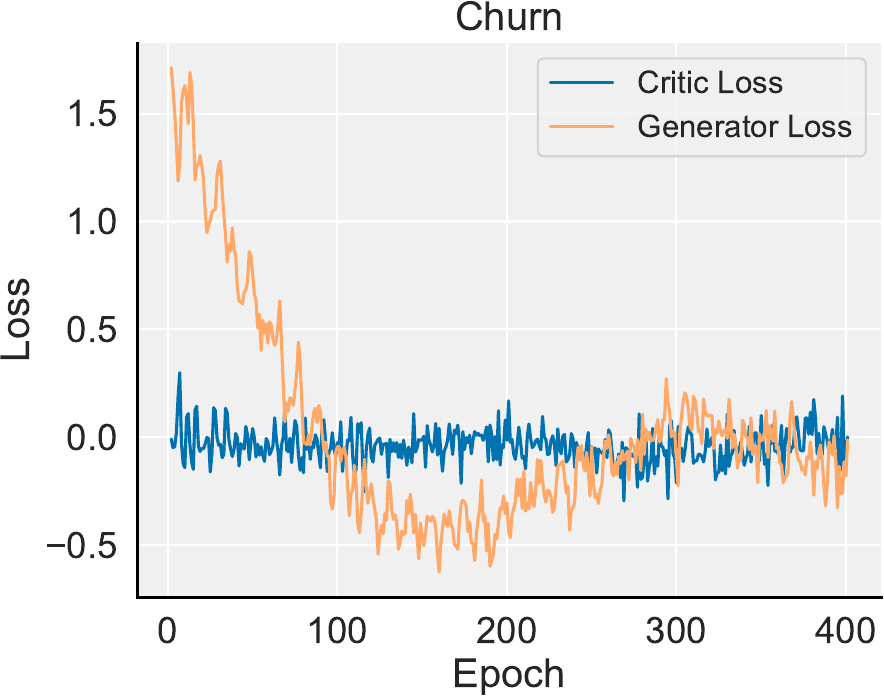}%
\includegraphics[width=.49\linewidth]{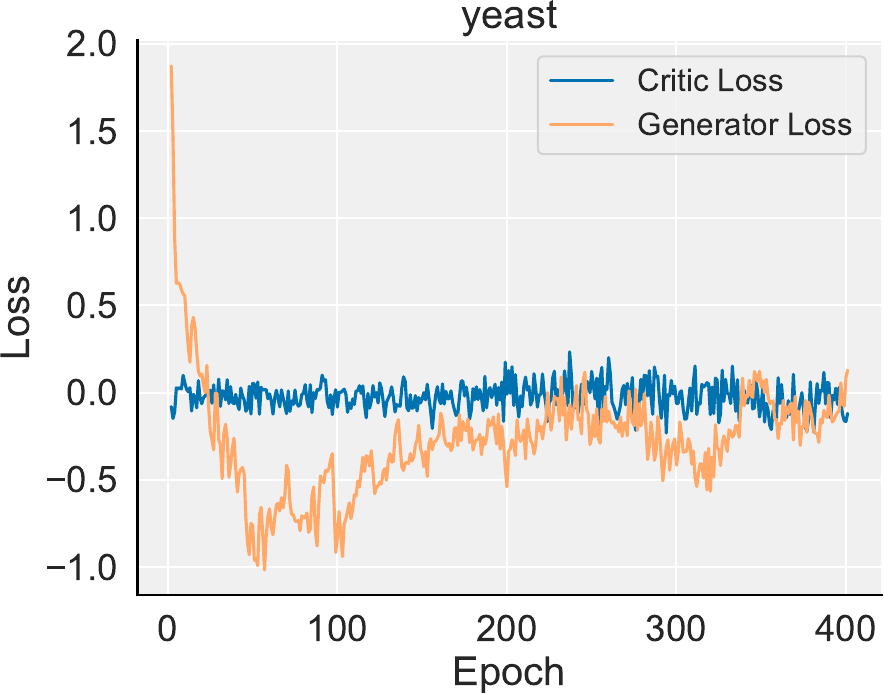}%\\\medskip

\caption{Plot of the $\loss_\critic$ and $\loss_\generator$ losses for the Churn Modelling
($m=10000$, $n_c=6$, $n_d=4$) and yeast ($m=1484$, $n_c=8$, $n_d=0$) datasets.}
\label{fig:loss}
\end{figure}

The joint introduction of $\loss_\mathbf{u}$ and $\loss_\mathbf{y}$ forces \ganname{} to learn not 
simply to produce class-specific samples, but also, to place these samples within appropriate
clusters. This approach leads to the synthesis of more realistic data. We clarify the term
``appropriate cluster'' in the next subsection.

From the perspective of stability, $\loss_\critic$ and $\loss_\generator$ exhibit remarkable
behavior. After convergence they fluctuate infinitesimally afterwards. We indicatively illustrate
two datasets in Fig.~\ref{fig:loss}.

\subsection{Probabilistic Sampling}
\label{ssec:train-samp}

Here we introduce a novel method applied by the Generator during the post-training generation
process (i.e., in the production phase of the model). We call it \textit{probabilistic sampling}
because it introduces additional conditions in a probabilistic manner with the aim of generating
more realistic samples.

The idea is as follows: when \ganname{} is requested to synthesize data instances belonging to a
specific class $y^*$, we try to create these samples in locations in the real space where it is
more possible to encounter that class. Therefore, in the original condition \textit{``create
samples of class $y=y^*$''}, we also consider the additional condition \textit{``create samples
inside clusters $u^*=\{u_1, u_2, \dots\}$''}, where $u_1, u_2, \dots$ are the clusters where it is
more possible to locate samples belonging to $y^*$.

To implement probabilistic sampling, we need to introduce the conditional probability
$P\left({U=u}\;\middle\vert\;{Y=y}\right)$ that quantifies the likelihood of a sample being placed
inside a cluster $u$, given its class $y$. The probability value can be easily computed during
training, by counting the $y$-class samples that simultaneously belong to cluster $u$:
\begin{equation}
\begin{aligned}
P\left({U=u}\;\middle\vert\;{Y=y}\right) &=
\frac{P\left({Y=y}\;\cap\;{U=u}\right)}{P\left(Y=y\right)} \\
 &=\frac{\vert\mathbf{X}\left\{\realx_i\;\middle\vert\;{y_i=y, u_i=u}\right\}\vert}
{\vert\mathbf{X}\left\{\realx_i\;\middle\vert\;{y_i=y}\right\}\vert}.
\end{aligned}
\end{equation}

Next, we define the following $|Y|\times k$ probability matrix:
\begin{equation}
\label{eq:pmatrix}
P_s=\left[\begin{smallmatrix}
P\left({U=u_1}\;\middle\vert\;{Y=y_1}\right) &\dots& P\left({U=u_k}\;\middle\vert\;{Y=y_1}\right)\\
P\left({U=u_1}\;\middle\vert\;{Y=y_2}\right)&\dots& P\left({U=u_k}\;\middle\vert\;{Y=y_2}\right)\\
 & & \\
     \svdots                                 &\svdots&        \svdots                            \\
 & & \\
P\left({U=u_1}\;\middle\vert\;{Y=y_{|Y|}}\right) &\dots& P\left({U=u_k}\;\middle\vert\;{Y=y_{|Y|}}\right) \\
\end{smallmatrix}
\right].
\end{equation}

Each row of $P^i_s$ corresponds to a class $y_i$ and stores the conditional probabilities that a
sample belonging to $y_i$ is also placed inside one of the $k$ clusters.

\begin{algorithm}[!t]
\caption{\ganname{} probabilistic sampling}
\label{algo:ps}
\hspace*{\algorithmicindent} \textbf{function} $\mathsf{sample}\big(\mathsf{N},\condd=\{\}\big)$ \\
   \hspace*{\algorithmicindent} \textbf{Input:} The number $\mathsf{N}$ of samples to be created
   and a set
   \hspace*{\algorithmicindent} \textbf{~~~~~~~~} of $\condd$ to be satisfied by the created
   samples. \\
   \hspace*{\algorithmicindent} \textbf{Output:} $\mathsf{N}$ synthetic samples.

\begin{algorithmic}[1]
\State $\mathbf{\hat{X}} \leftarrow \left[~\right]$
\While {$\vert\mathbf{\hat{X}}\vert < \mathsf{N}$}
	\State $\lat_{i}~\sim\mathcal{N}(\mu,\sigma)$
	\While {$i < n_d$}                                       \LineComment{For each discrete column}
		\If {$\condd[X^d_i]$ exists}
			\State{$\lat \leftarrow \lat \oplus h(\condd[X^d_i])$}     \LineComment{One-hot encode}
		\Else
			\State{$\lat \leftarrow \lat \oplus h(\randd(\{X^d_i\}))$} \LineComment{One-hot encode}
		\EndIf
	\EndWhile

	\If {$\condd[y]$ exists}                                          \LineComment{Class condition}
		\State {$\lat^y \leftarrow\condd[y]=h(y^*)$}
		\State {$\lat^u \leftarrow h(\randd(1,k,P_s^{y^*}))$} \LineComment{sample with $P_s^{y^*}$}
		\State{$\genx\leftarrow\generator(\lat \oplus \lat^u \oplus \lat^y)$}
		\If {$\geny == \lat^y$}
			$\mathbf{\hat{X}} \leftarrow \mathbf{\hat{X}} \cup \genx$
			\LineComment{append $\genx$ to $\mathbf{\hat{X}}$}
		\EndIf
	\Else                                                          \LineComment{no class condition}
		\State {$\lat^y \leftarrow h(\randd(1,|Y|))$}                \LineComment{sample uniformly}
		\State {$\lat^u \leftarrow\randd(1,k)$}                      \LineComment{sample uniformly}
		\State{$\genx\leftarrow\generator(\lat \oplus \lat^u \oplus \lat^y)$}
		\State{$\mathbf{\hat{X}} \leftarrow \mathbf{\hat{X}} \cup \genx$}
		\LineComment{append $\genx$ to $\mathbf{\hat{X}}$}
	\EndIf
\EndWhile
\State{\Return{$\mathbf{\hat{X}}$}}
\end{algorithmic}
\end{algorithm}

Algorithm~\ref{algo:ps} shows the basic steps for creating $\mathsf{N}$ samples under the
constraints declared in the $\condd$ dictionary. The expression $\condd[y]=y^*$ dictates that the
class of the generated samples should be $y^*$, whereas the expression $\condd[X^d_i]$ sets a
condition on the $X^d_i$ column. Steps 3--14 construct the latent vector $\lat$ according to
Eq.~\ref{eq:latsample}. In step 11, notice how a latent cluster is selected by accessing the row of
$P_s$ that corresponds to $y^*$.

As an example, consider the case where there are $|Y|=2$ classes, $k=3$ clusters, and
$P_s=\left[[0.7, 0.2, 0.1],[0.3, 0.7, 0]\right]$. In case a condition $\condd[y]=y_2$ is set, then,
we access the second row of $P_s$ to select either $u_1$ (with probability $0.3$), or $u_2$ (with
probability $0.7$). Reasonably, $u_3$ will never be selected, since no samples from class $y_2$
were originally placed in $u_3$ during the initial clustering phase.

Step 13 imposes the acceptance of samples that belong to the requested class $y^*$. This check is
crucial because there is no guarantee that the model will produce samples only from class $y^*$.
Accepting samples from other classes would actually intensify the problem of imbalance instead of
fixing it. It should be emphasized that the rejection step does not affect the conditional
probabilities $P(U \vert Y)$. These probabilities are estimated from the original data during 
training and remain fixed throughout the generation process.

Noticeably, the logic of Algorithm~\ref{algo:ps} contradicts several clustering-based oversampling
techniques that attempt to establish balance in a dataset by equalizing the number of samples in
each cluster (e.g. k-Means SMOTE~\cite{is2018b}). These techniques often fail to improve the
performance of downstream tasks because they generate samples in a way that is incompatible with 
the original data distribution (i.e., they create many samples in places where there are only a few
in the input dataset). In contrast, the proposed probabilistic sampling method respects the
original data distribution and this property is reflected in the experimental results presented
in Section \ref{sec:exp}.

\begin{table}[!t]
\begin{center}
\caption{Dataset characteristics: rows, columns, discrete and continuous columns, and number of classes.}
\label{tab:ds}
\begin{tabular}{| l | c | c | c | c | c |}            \hline
{\bf Dataset}                                        &  $m$  & $n$ & $n_d$ & $n_c$ & $|Y|$ \\\hline
Anemia\footnotemark[\getrefnumber{ds:kag}]           &  1281 & 14  &   0   &  14   &   9   \\
Churn Modelling\footnotemark[\getrefnumber{ds:kag}]  & 10000 & 10  &   4   &   6   &   2   \\
Dry Bean\footnotemark[\getrefnumber{ds:uci}]         & 13611 & 16  &   0   &  16   &   7   \\
ecoli1\footnotemark[\getrefnumber{ds:keel}]          &   336 &  7  &   0   &   7   &   2   \\
ecoli2\footnotemark[\getrefnumber{ds:keel}]          &   336 &  7  &   0   &   7   &   2   \\
ecoli4\footnotemark[\getrefnumber{ds:keel}]          &   336 &  7  &   0   &   7   &   2   \\
Fetal Health\footnotemark[\getrefnumber{ds:kag}]     &  2126 & 21  &   1   &  20   &   3   \\
Flare-F\footnotemark[\getrefnumber{ds:keel}]         &  1066 & 11  &   11  &   0   &   2   \\
glass5\footnotemark[\getrefnumber{ds:keel}]          &   214 &  9  &   0   &   9   &   2   \\
glass6\footnotemark[\getrefnumber{ds:keel}]          &   214 &  9  &   0   &   9   &   2   \\
Heart Disease\footnotemark[\getrefnumber{ds:uci}]    &   303 & 13  &   6   &   7   &   2   \\
New Thyroid\footnotemark[\getrefnumber{ds:keel}]     &   215 &  5  &   0   &   5   &   3   \\
yeast\footnotemark[\getrefnumber{ds:uci}]            &  1484 &  8  &   0   &   8   &  10   \\
yeast1\footnotemark[\getrefnumber{ds:uci}]           &  1484 &  8  &   0   &   8   &  2    \\\hline
\end{tabular}
\end{center}
\end{table}

\section{Experiments}
\label{sec:exp}

We evaluated \ganname{} in four tasks: i) improvement of classification performance, ii) generation 
of high fidelity synthetic data, iii) input memorization risk, and iv) privacy preservation. The
experiments have been conducted on a Linux Mint system with a Core i7-12700K CPU, 32GB of RAM and
NVIDIA RTX 3070 GPU. We have made the implementation of \ganname{} publicly available on
GitHub\footnote{https://github.com/lakritidis/ctdGAN}.

\subsection{Datasets and Models}
\label{ssec:ds}

This part of the study was designed to evaluate the performance of \ganname{} with reliable and
reproducible experiments. In this context, we utilized multiple well-established sources of tabular
datasets, such as Keel\footnote{\label{ds:keel}\url{https://sci2s.ugr.es/keel/imbalanced.php}}, 
Kaggle\footnote{\label{ds:kag}\url{https://www.kaggle.com/datasets}}, and the UCI Machine Learning
repository\footnote{\label{ds:uci}\url{https://archive.ics.uci.edu}}. From these sources, we
selected \numdatasets{} diverse datasets with the aim of covering multiple scenarios, like
binary/multiclass classification and combinations of numeric with categorical features. The
characteristics of these datasets are reported in Table \ref{tab:ds}.

The performance of \ganname{} was compared against numerous state-of-the-art generative models,
including ctGAN~\cite{anips2019}, CopulaGAN (COPGAN)~\cite{arxiv2021}, CTAB-GAN+~\cite{fbd2024},
SB-GAN \cite{ictai2023}, TVAE \cite{anips2019}, Conditional GAN (CGAN) \cite{arxiv2014} and
Gaussian Copula (GCOP) \cite{ejs2012}. For each of these models we retained all method-specific
components, including its preprocessing and normalization procedures, loss functions, conditioning
mechanism, sampling strategy and optimizer, as provided by the corresponding public implementation.
However, to improve the comparability of the experimental results, we harmonized a limited set of
general capacity and training-budget parameters that include the hidden-layer dimensions (256
neurons), latent dimensionality (128), number of training epochs (300), and batch sizes (100
samples). Table \ref{tab:hyperparams} illustrates the hyper-parameter values of the baseline
models.

We used the same hyper-parameters to train \ganname{}. The Critic, Generator and Classifier 
networks adopted the architecture of Subsection~\ref{ssec:arch}. They were trained by employing the
Adam optimizer with a learning rate equal to $\gamma=2\cdot 10^{-4}$ and a weight decay factor of
$10^{-6}$.

\begin{table*}[!t]
\begin{center}
\caption{Default, used and unchanged hyper-parameter values of the baseline models.}
\setlength\tabcolsep{4pt}
\label{tab:hyperparams}
\begin{tabular}{| l | l | l | c | c | l |}  \hline
\multicolumn{2}{|l|}{\bf Model} & {\bf Architecture} & {\bf Epochs} & {\bf Batch Size} & {\bf Default components (unchanged)} \\\hline

\multirow{2}{*}{ctGAN} & Default & $\generator{}(256,256), \critic{}(256,256)$ & $300$ & $500$ & \multirow{2}{9.4cm}{Loss function, $pac=10$, Data normalization (VGM, One Hot), Adam optimizer, Learning rate: $2\cdot 10^{-4}$, Weight decay: $10^{-6}$, Latent vector dimensions $e=128$.} \\
      & Used    & $\generator{}(256,256), \critic{}(256,256)$ & $300$ & $100$ & \\\hline

Copula & Default & $\generator{}(256,256), \critic{}(256,256)$ & $300$ & $500$ & \multirow{2}{9.4cm}{Loss function, $pac=10$, Data normalization (VGM, One Hot), Adam optimizer, Learning rate: $2\cdot 10^{-4}$, Weight decay: $10^{-6}$, Latent vector dimensions $e=128$.} \\
GAN   & Used    & $\generator{}(256,256), \critic{}(256,256)$ & $300$ & $100$ & \\\hline

CTAB- & Default & $\generator{}(256,256), \critic{}(256,256)$ & $100$ & $500$ & \multirow{3}{9.4cm}{Loss function, $pac=10$, Normalization (VGM, One Hot, dedicated encoders), Internal classifier architecture, Adam optimizer, Learning rate: $2\cdot 10^{-4}$, Weight decay: $10^{-6}$, Latent vector dimensions $e=128$.} \\
GAN+   & Used    & $\generator{}(256,256), \critic{}(256,256)$ & $300$ & $100$ & \\
       &         &  &  &  & \\\hline

Cond. & Default & $\generator{}(256,256), \critic{}(256,256)$ & $300$ & $32$ & \multirow{2}{9.4cm}{Loss function, Normalization, Adam optimizer, Learning rate: $2\cdot 10^{-4}$, Weight decay: $10^{-6}$, Latent vector dimensions $e=128$.} \\
GAN   & Used    & $\generator{}(256,256), \critic{}(256,256)$ & $300$ & $100$ & \\\hline

SB- & Default & $\generator{}(64,32), \critic{}(64,32)$ & $100$ & $32$ & \multirow{2}{9.4cm}{Loss function, Normalization, Adam optimizer, Learning rate: $2\cdot 10^{-4}$, Weight decay: $10^{-6}$, Latent vector dimensions $e=8$.} \\
GAN   & Used    & $\generator{}(256,256), \critic{}(256,256)$ & $300$ & $100$ & \\\hline

\multirow{2}{*}{TVAE} & Default & $(128,128), (128,128)$ & $300$ & $500$ & \multirow{2}{9.4cm}{Loss function, Data normalization (VGM, One Hot), Adam optimizer, Learning rate: $2\cdot 10^{-4}$, Weight decay: $10^{-6}$, Latent vector dimensions $e=128$.} \\
      & Used    & $(256,256), (256,256)$ & $300$ & $100$ & \\\hline
      
%\makecell{TVAE} & \makecell{Encoder - Decoder\\(128,128) - (128,128)\\(256,256) - (256,256)} & \makecell{300\\(300)} & \makecell{500\\(100)} & \makecell{Loss function, \\Normaliz. (VGM - One\\Hot), Adam optimizer,\\Learning rate: $\cdot 10^{-3}$,\\Weight decay: $10^{-5}$,\\Lat. vect. dim. $e=128$}  \\\hline

%\makecell{Gauss.\\Copula} & \makecell{$\generator{}$(256,256), $\critic{}$(256,256)\\$\generator{}$(, ),$\critic{}$ (, )} & \makecell{300\\( )} & \makecell{500\\(?)} & \makecell{Loss function,\\Normal. (VGM, OHE),\\Adam optimizer,\\Learning rate: $10^{-4}$,\\Weight decay $10^{-6}$} \\\hline
\end{tabular}
\end{center}
\end{table*}

To mitigate the effects of randomness and enhance statistical significance, we repeated all the 
experiments \numseeds{} times, by using different random states (i.e., 0, 1, and 42). The presented 
results constitute the average values over these \numseeds{} executions. Their statistical
significance was verified by using the Friedman non-parametric test. The obtained $p$-values are
presented in the top of each result table, and in all cases, they were lower than $5\cdot10^{-2}$,
leading to the safe rejection of the null hypothesis.

\subsection{Improvement of Classification Performance}
\label{ssec:exp-over}

This experiment evaluates the ability of a generative model $\mathcal{M}$ to improve classification
performance. In this test we also included three popular oversampling techniques: SMOTE
\cite{jair2002}, ADASYN \cite{ijcnn2008} and k-Means SMOTE \cite{is2018b}. Initially, $\mathcal{M}$
synthesizes minority samples in order to nullify the imbalance in a portion of the input dataset. 
Next, a classifier $\mathcal{P}$ is trained on the balanced portion and tested on the imbalanced
one.

The classification performance was evaluated by using two well-established measures: the $F1$ score
and Balanced Accuracy $B_{ac}$. The former is defined as the harmonic mean of Precision $P$ and
Recall $R$, namely, $F1={2PR}/{(P+R)}$. In all cases, we used the macro $F1$ variant that computes
the $F1$ score for each class individually and returns their average value. On the other hand,
Balanced Accuracy is the average of Recall obtained on each involved class. It is one of the most
common evaluation measures in both binary and multi-class classification tasks with imbalanced
data.

\begin{figure}[!b]
\center
\includegraphics[width=0.95\linewidth]{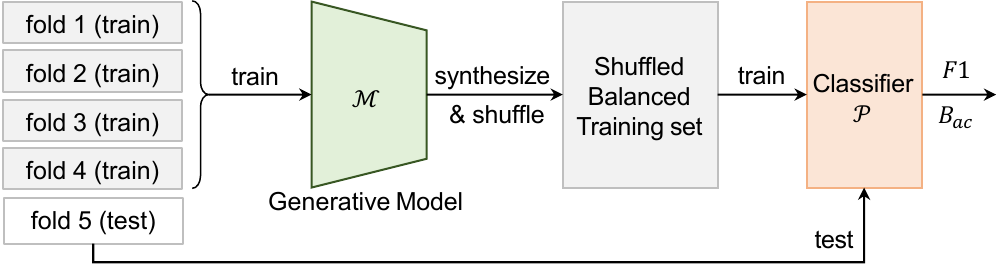}
\caption{Experiment 1: Improving the performance of a classifier $\mathcal{P}$ by oversampling with 
a generative model $\mathcal{M}$.}
\label{fig:exp-os}
\bigskip
\includegraphics[width=0.95\linewidth]{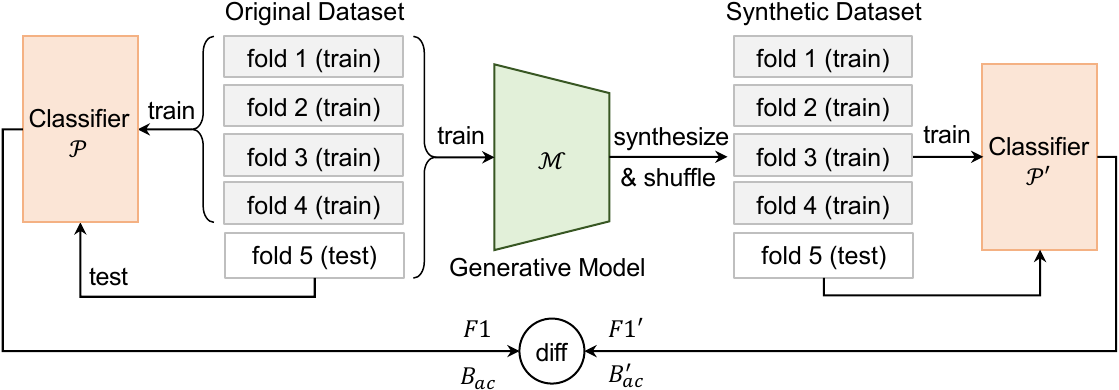}
\caption{The stages of the data fidelity (ML utility) experiment.}
\label{fig:exp-fid}
\end{figure}

The values of $F1$ and $B_{ac}$ were validated by applying 5-fold cross validation, as illustrated
in Fig. \ref{fig:exp-os}. At each CV iteration, $\mathcal{M}$ was trained on 4 out of the 5 folds;
then, it was used to establish balance in these 4 folds by synthesizing the necessary number of
samples from each minority class. The classifier $\mathcal{P}$ was subsequently trained on the
four balanced folds, and tested on the fifth one. As mentioned earlier, the clustering step and the
pre-training phase of the two auxiliary classifiers $\clfqy$ and $\clfqu$ take place inside the
training process of \ganname{} using the same training examples, to prevent information leakage
from the training to the test set.

Regarding $\mathcal{P}$, we employed three predictors: i) XGBoost (XGB), which is among the
strongest tabular data classifiers, ii) a Multilayer Perceptron (MLP) with two hidden layers of 128
neurons and $\relu$ activation, and iii) a Random Forest (RF) with 50 estimators and unbounded
maximum tree depth.

Table~\ref{tab:os} illustrates the $F1$ score for XGBoost and $B_{ac}$ for all 3 classifiers. The
measure values concern the mean (over \numseeds{} random states) of the mean (over 5 CV folds)
$F1$ and $B_{ac}$ values achieved by the classifiers on datasets that have been balanced with the
involved models. The last rows of the sub-tables reveal the mean ranking of each model, whereas the
parenthesized numbers denote the number of first, second, and last positions in the rankings.

% -----------------------------------------------------------------------------------------------
% -----------------------------------------------------------------------------------------------
% CLASSIFICATION PERFORMANCE (OVERSAMPLING) EXPERIMENT 
% -----------------------------------------------------------------------------------------------
% -----------------------------------------------------------------------------------------------
\begin{table*}
\begin{center}
\caption{
Classification performance of XGBoost, MLP and Random Forest on \numdatasets{} datasets balanced
with various generative models.}
\label{tab:os}
\setlength\tabcolsep{1.298pt}

% -----------------------------------------------------------------------------------------------
% XGBOOST - F1
% -----------------------------------------------------------------------------------------------
\begin{tabular}{|l||P{1.4cm}|P{1.4cm}|P{1.6cm}|P{1.4cm}|P{1.4cm}|P{1.5cm}|P{1.4cm}|P{1.4cm}|P{1.4cm}|P{1.4cm}|P{1.5cm}|}

\multicolumn{12}{c}{\bf XGBoost Classifier -- $F1$ score -- Friedman test $p$-value: $1.150941\cdot 10^{-3}$}\\\hline
{\bf Dataset} & {\bf SMOTE} & {\bf ADASYN} & {\bf KM-SMOTE} & {\bf CGAN} & {\bf COPGAN} & {\bf CTAB-GAN} & {\bf CTGAN} & {\bf GCOP} & {\bf SB-GAN} & {\bf TVAE} & {\bf \ganname{}}  \\\hline

Anemia  &  0.950$\pm$\scriptsize{0.02} & \underline{0.955$\pm$\scriptsize{0.02}} & \underline{\textbf{0.955$\pm$\scriptsize{0.02}}} & 0.942$\pm$\scriptsize{0.02} & 0.867$\pm$\scriptsize{0.02} & 0.856$\pm$\scriptsize{0.03} & 0.886$\pm$\scriptsize{0.03} & 0.884$\pm$\scriptsize{0.02} & 0.943$\pm$\scriptsize{0.02} & 0.867$\pm$\scriptsize{0.02} & 0.919$\pm$\scriptsize{0.02}  \\
Churn  &  \underline{\textbf{0.595$\pm$\scriptsize{0.01}}} & \underline{0.593$\pm$\scriptsize{0.01}} & 0.585$\pm$\scriptsize{0.01} & 0.586$\pm$\scriptsize{0.01} & 0.497$\pm$\scriptsize{0.01} & 0.583$\pm$\scriptsize{0.01} & 0.539$\pm$\scriptsize{0.01} & 0.556$\pm$\scriptsize{0.01} & 0.579$\pm$\scriptsize{0.01} & 0.580$\pm$\scriptsize{0.01} & 0.587$\pm$\scriptsize{0.01}  \\
Dry B.  &  0.938$\pm$\scriptsize{0.00} & 0.937$\pm$\scriptsize{0.00} & 0.938$\pm$\scriptsize{0.00} & 0.938$\pm$\scriptsize{0.00} & 0.937$\pm$\scriptsize{0.00} & 0.936$\pm$\scriptsize{0.00} & 0.936$\pm$\scriptsize{0.00} & 0.936$\pm$\scriptsize{0.00} & \underline{0.940$\pm$\scriptsize{0.00}} & 0.935$\pm$\scriptsize{0.00} & \underline{\textbf{0.941$\pm$\scriptsize{0.00}}}  \\
ecoli1  &  0.784$\pm$\scriptsize{0.04} & \underline{\textbf{0.808$\pm$\scriptsize{0.04}}} & 0.781$\pm$\scriptsize{0.05} & 0.786$\pm$\scriptsize{0.05} & 0.785$\pm$\scriptsize{0.04} & 0.778$\pm$\scriptsize{0.05} & \underline{0.800$\pm$\scriptsize{0.05}} & 0.782$\pm$\scriptsize{0.05} & 0.779$\pm$\scriptsize{0.05} & 0.790$\pm$\scriptsize{0.05} & 0.787$\pm$\scriptsize{0.06}  \\
ecoli2  &  0.804$\pm$\scriptsize{0.06} & 0.781$\pm$\scriptsize{0.06} & 0.798$\pm$\scriptsize{0.06} & 0.785$\pm$\scriptsize{0.07} & 0.808$\pm$\scriptsize{0.06} & 0.801$\pm$\scriptsize{0.07} & \underline{0.809$\pm$\scriptsize{0.07}} & 0.792$\pm$\scriptsize{0.07} & \underline{\textbf{0.820$\pm$\scriptsize{0.05}}} & 0.799$\pm$\scriptsize{0.05} & 0.785$\pm$\scriptsize{0.06}  \\
ecoli4  &  \underline{0.795$\pm$\scriptsize{0.06}} & 0.771$\pm$\scriptsize{0.06} & 0.776$\pm$\scriptsize{0.09} & 0.776$\pm$\scriptsize{0.09} & 0.763$\pm$\scriptsize{0.08} & 0.766$\pm$\scriptsize{0.07} & 0.762$\pm$\scriptsize{0.07} & 0.737$\pm$\scriptsize{0.12} & 0.781$\pm$\scriptsize{0.07} & 0.793$\pm$\scriptsize{0.07} & \underline{\textbf{0.796$\pm$\scriptsize{0.06}}}  \\
Fetal  &  0.909$\pm$\scriptsize{0.01} & 0.901$\pm$\scriptsize{0.01} & 0.902$\pm$\scriptsize{0.01} & 0.901$\pm$\scriptsize{0.01} & \underline{0.913$\pm$\scriptsize{0.01}} & 0.910$\pm$\scriptsize{0.01} & 0.908$\pm$\scriptsize{0.01} & 0.912$\pm$\scriptsize{0.01} & 0.909$\pm$\scriptsize{0.01} & 0.899$\pm$\scriptsize{0.01} & \underline{\textbf{0.914$\pm$\scriptsize{0.01}}}  \\
flare  &  0.218$\pm$\scriptsize{0.05} & 0.220$\pm$\scriptsize{0.04} & 0.167$\pm$\scriptsize{0.06} & 0.156$\pm$\scriptsize{0.03} & 0.265$\pm$\scriptsize{0.04} & 0.144$\pm$\scriptsize{0.07} & 0.265$\pm$\scriptsize{0.04} & 0.103$\pm$\scriptsize{0.02} & 0.170$\pm$\scriptsize{0.03} & \underline{\textbf{0.303$\pm$\scriptsize{0.05}}} & \underline{0.267$\pm$\scriptsize{0.03}}  \\
glass5  &  \underline{0.706$\pm$\scriptsize{0.17}} & 0.687$\pm$\scriptsize{0.19} & 0.656$\pm$\scriptsize{0.21} & 0.569$\pm$\scriptsize{0.17} & 0.684$\pm$\scriptsize{0.15} & 0.638$\pm$\scriptsize{0.13} & 0.693$\pm$\scriptsize{0.13} & 0.591$\pm$\scriptsize{0.15} & 0.576$\pm$\scriptsize{0.17} & 0.620$\pm$\scriptsize{0.20} & \underline{\textbf{0.759$\pm$\scriptsize{0.09}}}  \\
glass6  &  0.870$\pm$\scriptsize{0.03} & 0.844$\pm$\scriptsize{0.06} & 0.851$\pm$\scriptsize{0.04} & \underline{\textbf{0.894$\pm$\scriptsize{0.05}}} & 0.823$\pm$\scriptsize{0.05} & 0.833$\pm$\scriptsize{0.05} & 0.852$\pm$\scriptsize{0.05} & 0.850$\pm$\scriptsize{0.03} & \underline{0.877$\pm$\scriptsize{0.05}} & 0.833$\pm$\scriptsize{0.05} & 0.866$\pm$\scriptsize{0.04}  \\
Heart D. &  0.813$\pm$\scriptsize{0.02} & \underline{\textbf{0.827$\pm$\scriptsize{0.02}}} & 0.819$\pm$\scriptsize{0.02} & 0.813$\pm$\scriptsize{0.02} & 0.810$\pm$\scriptsize{0.02} & 0.822$\pm$\scriptsize{0.02} & 0.809$\pm$\scriptsize{0.03} & 0.815$\pm$\scriptsize{0.02} & 0.818$\pm$\scriptsize{0.02} & 0.811$\pm$\scriptsize{0.02} & \underline{0.823$\pm$\scriptsize{0.01}}  \\
New T.  &  0.918$\pm$\scriptsize{0.03} & 0.922$\pm$\scriptsize{0.03} & 0.931$\pm$\scriptsize{0.02} & \underline{0.934$\pm$\scriptsize{0.04}} & 0.921$\pm$\scriptsize{0.03} & 0.847$\pm$\scriptsize{0.03} & 0.899$\pm$\scriptsize{0.03} & 0.838$\pm$\scriptsize{0.04} & \underline{\textbf{0.945$\pm$\scriptsize{0.04}}} & 0.844$\pm$\scriptsize{0.03} & 0.896$\pm$\scriptsize{0.02}  \\
yeast  &  \underline{0.537$\pm$\scriptsize{0.02}} & 0.537$\pm$\scriptsize{0.02} & 0.537$\pm$\scriptsize{0.02} & 0.532$\pm$\scriptsize{0.03} & 0.502$\pm$\scriptsize{0.04} & 0.505$\pm$\scriptsize{0.03} & 0.504$\pm$\scriptsize{0.03} & 0.520$\pm$\scriptsize{0.03} & 0.518$\pm$\scriptsize{0.03} & 0.533$\pm$\scriptsize{0.03} & \underline{\textbf{0.550$\pm$\scriptsize{0.03}}}  \\
yeast1  &  0.488$\pm$\scriptsize{0.08} & 0.483$\pm$\scriptsize{0.07} & 0.502$\pm$\scriptsize{0.09} & 0.477$\pm$\scriptsize{0.06} & 0.490$\pm$\scriptsize{0.09} & 0.510$\pm$\scriptsize{0.07} & 0.478$\pm$\scriptsize{0.09} & 0.479$\pm$\scriptsize{0.06} & \underline{0.513$\pm$\scriptsize{0.06}} & 0.492$\pm$\scriptsize{0.07} & \underline{\textbf{0.516$\pm$\scriptsize{0.07}}}  
\\\hline\hline

{\bf Rank} & \underline{4.50 (1-3-0)} & 5.04 (2-2-1) & 5.25 (1-0-0) & 6.29 (1-1-2) & 7.11 (0-1-3) & 7.50 (0-0-2) & 6.82 (0-2-1) & 8.36 (0-0-3) & 4.93 (2-3-0) & 6.93 (1-0-2) & \underline{\textbf{3.29 (6-2-0)}} \\\hline
\end{tabular}

\medskip

% -----------------------------------------------------------------------------------------------
% XGBOOST - BACC
% -----------------------------------------------------------------------------------------------
\begin{tabular}{|l||P{1.4cm}|P{1.4cm}|P{1.6cm}|P{1.4cm}|P{1.4cm}|P{1.5cm}|P{1.4cm}|P{1.4cm}|P{1.4cm}|P{1.4cm}|P{1.5cm}|}

\multicolumn{12}{c}{\bf XGBoost Classifier -- Balanced Accuracy -- Friedman test $p$-value: $4.834400\cdot 10^{-6}$}\\\hline
{\bf Dataset} & {\bf SMOTE} & {\bf ADASYN} & {\bf KM-SMOTE} & {\bf CGAN} & {\bf COPGAN} & {\bf CTAB-GAN} & {\bf CTGAN} & {\bf GCOP} & {\bf SB-GAN} & {\bf TVAE} & {\bf \ganname{}}  \\\hline

Anemia  &  \underline{\textbf{0.957$\pm$\scriptsize{0.02}}} & \underline{0.956$\pm$\scriptsize{0.02}} & 0.956$\pm$\scriptsize{0.02} & 0.910$\pm$\scriptsize{0.03} & 0.892$\pm$\scriptsize{0.02} & 0.887$\pm$\scriptsize{0.02} & 0.915$\pm$\scriptsize{0.02} & 0.902$\pm$\scriptsize{0.02} & 0.920$\pm$\scriptsize{0.02} & 0.909$\pm$\scriptsize{0.02} & 0.932$\pm$\scriptsize{0.02}  \\
Churn  &  \underline{\textbf{0.734$\pm$\scriptsize{0.01}}} & 0.732$\pm$\scriptsize{0.01} & 0.723$\pm$\scriptsize{0.01} & 0.720$\pm$\scriptsize{0.01} & 0.692$\pm$\scriptsize{0.01} & 0.722$\pm$\scriptsize{0.01} & 0.726$\pm$\scriptsize{0.01} & 0.717$\pm$\scriptsize{0.01} & 0.724$\pm$\scriptsize{0.01} & \underline{0.732$\pm$\scriptsize{0.01}} & 0.724$\pm$\scriptsize{0.01}  \\
Dry B.  &  0.938$\pm$\scriptsize{0.00} & 0.938$\pm$\scriptsize{0.00} & \underline{0.938$\pm$\scriptsize{0.00}} & 0.935$\pm$\scriptsize{0.00} & 0.936$\pm$\scriptsize{0.00} & 0.936$\pm$\scriptsize{0.00} & 0.936$\pm$\scriptsize{0.00} & 0.935$\pm$\scriptsize{0.00} & 0.935$\pm$\scriptsize{0.00} & 0.934$\pm$\scriptsize{0.00} & \underline{\textbf{0.938$\pm$\scriptsize{0.00}}}  \\
ecoli1  &  0.868$\pm$\scriptsize{0.02} & \underline{\textbf{0.887$\pm$\scriptsize{0.02}}} & 0.862$\pm$\scriptsize{0.03} & 0.869$\pm$\scriptsize{0.03} & 0.865$\pm$\scriptsize{0.02} & 0.857$\pm$\scriptsize{0.03} & 0.868$\pm$\scriptsize{0.03} & 0.864$\pm$\scriptsize{0.03} & \underline{0.876$\pm$\scriptsize{0.02}} & 0.871$\pm$\scriptsize{0.03} & 0.874$\pm$\scriptsize{0.04}  \\
ecoli2  &  0.885$\pm$\scriptsize{0.03} & 0.884$\pm$\scriptsize{0.03} & 0.870$\pm$\scriptsize{0.03} & \underline{\textbf{0.895$\pm$\scriptsize{0.03}}} & 0.867$\pm$\scriptsize{0.03} & 0.867$\pm$\scriptsize{0.04} & 0.874$\pm$\scriptsize{0.04} & 0.865$\pm$\scriptsize{0.04} & 0.877$\pm$\scriptsize{0.03} & 0.887$\pm$\scriptsize{0.03} & \underline{0.890$\pm$\scriptsize{0.04}}  \\
ecoli4  &  0.898$\pm$\scriptsize{0.04} & 0.885$\pm$\scriptsize{0.03} & 0.877$\pm$\scriptsize{0.05} & 0.889$\pm$\scriptsize{0.04} & 0.869$\pm$\scriptsize{0.04} & 0.859$\pm$\scriptsize{0.04} & 0.890$\pm$\scriptsize{0.04} & 0.871$\pm$\scriptsize{0.05} & 0.894$\pm$\scriptsize{0.03} & \underline{0.898$\pm$\scriptsize{0.04}} & \underline{\textbf{0.923$\pm$\scriptsize{0.04}}}  \\
Fetal H. &  \underline{\textbf{0.906$\pm$\scriptsize{0.01}}} & 0.901$\pm$\scriptsize{0.01} & 0.899$\pm$\scriptsize{0.01} & 0.899$\pm$\scriptsize{0.01} & \underline{0.904$\pm$\scriptsize{0.01}} & 0.902$\pm$\scriptsize{0.01} & 0.899$\pm$\scriptsize{0.01} & 0.903$\pm$\scriptsize{0.01} & 0.899$\pm$\scriptsize{0.01} & 0.895$\pm$\scriptsize{0.01} & 0.903$\pm$\scriptsize{0.01}  \\
Flare-F  &  0.640$\pm$\scriptsize{0.04} & 0.633$\pm$\scriptsize{0.03} & 0.553$\pm$\scriptsize{0.03} & 0.569$\pm$\scriptsize{0.05} & 0.720$\pm$\scriptsize{0.05} & 0.547$\pm$\scriptsize{0.03} & 0.720$\pm$\scriptsize{0.05} & 0.563$\pm$\scriptsize{0.04} & 0.580$\pm$\scriptsize{0.06} & \underline{0.740$\pm$\scriptsize{0.04}} & \underline{\textbf{0.758$\pm$\scriptsize{0.03}}}  \\
glass5  &  0.841$\pm$\scriptsize{0.08} & 0.829$\pm$\scriptsize{0.09} & 0.782$\pm$\scriptsize{0.09} & 0.766$\pm$\scriptsize{0.08} & 0.841$\pm$\scriptsize{0.08} & 0.847$\pm$\scriptsize{0.08} & \underline{0.873$\pm$\scriptsize{0.07}} & 0.856$\pm$\scriptsize{0.07} & 0.766$\pm$\scriptsize{0.08} & 0.793$\pm$\scriptsize{0.09} & \underline{\textbf{0.890$\pm$\scriptsize{0.08}}}  \\
glass6  &  \underline{0.933$\pm$\scriptsize{0.03}} & 0.923$\pm$\scriptsize{0.03} & 0.917$\pm$\scriptsize{0.03} & 0.929$\pm$\scriptsize{0.03} & 0.910$\pm$\scriptsize{0.03} & 0.922$\pm$\scriptsize{0.03} & 0.924$\pm$\scriptsize{0.03} & 0.924$\pm$\scriptsize{0.03} & 0.920$\pm$\scriptsize{0.03} & 0.912$\pm$\scriptsize{0.03} & \underline{\textbf{0.934$\pm$\scriptsize{0.03}}}  \\
Heart D. &  0.792$\pm$\scriptsize{0.02} & \underline{\textbf{0.805$\pm$\scriptsize{0.02}}} & 0.799$\pm$\scriptsize{0.02} & 0.803$\pm$\scriptsize{0.03} & 0.801$\pm$\scriptsize{0.02} & 0.801$\pm$\scriptsize{0.02} & 0.798$\pm$\scriptsize{0.02} & 0.797$\pm$\scriptsize{0.02} & 0.802$\pm$\scriptsize{0.02} & 0.796$\pm$\scriptsize{0.02} & \underline{0.805$\pm$\scriptsize{0.02}}  \\
N. Thyr.  &  0.920$\pm$\scriptsize{0.03} & 0.924$\pm$\scriptsize{0.03} & 0.920$\pm$\scriptsize{0.03} & 0.912$\pm$\scriptsize{0.03} & \underline{\textbf{0.930$\pm$\scriptsize{0.03}}} & 0.910$\pm$\scriptsize{0.02} & 0.916$\pm$\scriptsize{0.03} & 0.889$\pm$\scriptsize{0.03} & 0.925$\pm$\scriptsize{0.02} & \underline{0.929$\pm$\scriptsize{0.01}} & 0.925$\pm$\scriptsize{0.02}  \\
yeast  &  0.531$\pm$\scriptsize{0.02} & 0.531$\pm$\scriptsize{0.02} & 0.531$\pm$\scriptsize{0.02} & \underline{0.544$\pm$\scriptsize{0.02}} & 0.509$\pm$\scriptsize{0.03} & 0.510$\pm$\scriptsize{0.03} & 0.503$\pm$\scriptsize{0.03} & 0.524$\pm$\scriptsize{0.03} & 0.529$\pm$\scriptsize{0.02} & 0.543$\pm$\scriptsize{0.03} & \underline{\textbf{0.556$\pm$\scriptsize{0.03}}}  \\
yeast1  &  \underline{0.772$\pm$\scriptsize{0.05}} & \underline{\textbf{0.796$\pm$\scriptsize{0.05}}} & 0.707$\pm$\scriptsize{0.05} & 0.717$\pm$\scriptsize{0.05} & 0.703$\pm$\scriptsize{0.05} & 0.711$\pm$\scriptsize{0.05} & 0.717$\pm$\scriptsize{0.05} & 0.703$\pm$\scriptsize{0.04} & 0.725$\pm$\scriptsize{0.05} & 0.727$\pm$\scriptsize{0.04} & 0.729$\pm$\scriptsize{0.05}  \\\hline\hline

{\bf Rank} & \underline{4.07 (3-2-1)} & 4.25 (3-1-0) & 7.32 (0-1-0) & 6.14 (1-1-1) & 7.39 (1-1-2) & 8.21 (0-0-3) & 6.32 (0-1-1) & 8.21 (0-0-4) & 6.00 (0-1-0) & 5.71 (0-4-2) & \underline{\textbf{2.36 (6-2-0)}}  \\\hline
\end{tabular}

\bigskip

% -----------------------------------------------------------------------------------------------
% MLP - BACC
% -----------------------------------------------------------------------------------------------
\begin{tabular}{|l||P{1.4cm}|P{1.4cm}|P{1.6cm}|P{1.4cm}|P{1.4cm}|P{1.5cm}|P{1.4cm}|P{1.4cm}|P{1.4cm}|P{1.4cm}|P{1.5cm}|}

\multicolumn{12}{c}{\bf Multilayer Perceptron Classifier -- Balanced Accuracy -- Friedman test $p$-value: $7.958514\cdot 10^{-4}$}\\\hline
{\bf Dataset} & {\bf SMOTE} & {\bf ADASYN} & {\bf KM-SMOTE} & {\bf CGAN} & {\bf COPGAN} & {\bf CTAB-GAN} & {\bf CTGAN} & {\bf GCOP} & {\bf SB-GAN} & {\bf TVAE} & {\bf \ganname{}}  \\\hline

Anemia  &  \underline{\textbf{0.828$\pm$\scriptsize{0.02}}} & \underline{0.823$\pm$\scriptsize{0.02}} & 0.823$\pm$\scriptsize{0.02} & 0.738$\pm$\scriptsize{0.03} & 0.751$\pm$\scriptsize{0.03} & 0.710$\pm$\scriptsize{0.03} & 0.762$\pm$\scriptsize{0.03} & 0.694$\pm$\scriptsize{0.03} & 0.742$\pm$\scriptsize{0.03} & 0.802$\pm$\scriptsize{0.02} & 0.758$\pm$\scriptsize{0.03}  \\
Churn  &  \underline{0.719$\pm$\scriptsize{0.01}} & \underline{\textbf{0.721$\pm$\scriptsize{0.01}}} & 0.709$\pm$\scriptsize{0.01} & 0.703$\pm$\scriptsize{0.01} & 0.672$\pm$\scriptsize{0.01} & 0.702$\pm$\scriptsize{0.01} & 0.700$\pm$\scriptsize{0.01} & 0.690$\pm$\scriptsize{0.01} & 0.703$\pm$\scriptsize{0.01} & 0.710$\pm$\scriptsize{0.01} & 0.715$\pm$\scriptsize{0.01}  \\
Dry B.  &  0.935$\pm$\scriptsize{0.00} & 0.934$\pm$\scriptsize{0.00} & \underline{0.935$\pm$\scriptsize{0.00}} & 0.934$\pm$\scriptsize{0.00} & 0.934$\pm$\scriptsize{0.00} & 0.934$\pm$\scriptsize{0.00} & 0.935$\pm$\scriptsize{0.00} & 0.934$\pm$\scriptsize{0.00} & 0.932$\pm$\scriptsize{0.00} & 0.933$\pm$\scriptsize{0.00} & \underline{\textbf{0.936$\pm$\scriptsize{0.00}}}  \\
ecoli1  &  0.873$\pm$\scriptsize{0.03} & 0.862$\pm$\scriptsize{0.03} & 0.862$\pm$\scriptsize{0.03} & 0.858$\pm$\scriptsize{0.03} & 0.857$\pm$\scriptsize{0.03} & 0.869$\pm$\scriptsize{0.04} & 0.857$\pm$\scriptsize{0.04} & 0.846$\pm$\scriptsize{0.03} & 0.862$\pm$\scriptsize{0.03} & \underline{0.879$\pm$\scriptsize{0.03}} & \underline{\textbf{0.882$\pm$\scriptsize{0.03}}}  \\
ecoli2  &  0.883$\pm$\scriptsize{0.04} & 0.884$\pm$\scriptsize{0.03} & 0.878$\pm$\scriptsize{0.03} & 0.881$\pm$\scriptsize{0.03} & 0.888$\pm$\scriptsize{0.03} & 0.885$\pm$\scriptsize{0.02} & 0.863$\pm$\scriptsize{0.04} & 0.876$\pm$\scriptsize{0.04} & 0.881$\pm$\scriptsize{0.03} & \underline{\textbf{0.915$\pm$\scriptsize{0.02}}} & \underline{0.899$\pm$\scriptsize{0.02}}  \\
ecoli4  &  0.894$\pm$\scriptsize{0.03} & 0.892$\pm$\scriptsize{0.03} & 0.896$\pm$\scriptsize{0.03} & 0.904$\pm$\scriptsize{0.04} & 0.914$\pm$\scriptsize{0.03} & 0.919$\pm$\scriptsize{0.03} & \underline{\textbf{0.922$\pm$\scriptsize{0.03}}} & 0.910$\pm$\scriptsize{0.03} & 0.904$\pm$\scriptsize{0.04} & \underline{0.919$\pm$\scriptsize{0.04}} & 0.915$\pm$\scriptsize{0.04}  \\
Fetal H. &  \underline{0.873$\pm$\scriptsize{0.01}} & \underline{\textbf{0.875$\pm$\scriptsize{0.01}}} & 0.872$\pm$\scriptsize{0.01} & 0.834$\pm$\scriptsize{0.01} & 0.861$\pm$\scriptsize{0.02} & 0.854$\pm$\scriptsize{0.01} & 0.866$\pm$\scriptsize{0.01} & 0.853$\pm$\scriptsize{0.01} & 0.832$\pm$\scriptsize{0.01} & 0.859$\pm$\scriptsize{0.02} & 0.861$\pm$\scriptsize{0.01}  \\
Flare-F  &  0.618$\pm$\scriptsize{0.04} & 0.616$\pm$\scriptsize{0.04} & 0.550$\pm$\scriptsize{0.02} & 0.573$\pm$\scriptsize{0.05} & \underline{0.730$\pm$\scriptsize{0.05}} & 0.618$\pm$\scriptsize{0.05} & 0.730$\pm$\scriptsize{0.05} & 0.572$\pm$\scriptsize{0.04} & 0.572$\pm$\scriptsize{0.06} & 0.722$\pm$\scriptsize{0.05} & \underline{\textbf{0.755$\pm$\scriptsize{0.04}}}  \\
glass5  &  0.896$\pm$\scriptsize{0.07} & 0.909$\pm$\scriptsize{0.07} & 0.847$\pm$\scriptsize{0.07} & \underline{0.936$\pm$\scriptsize{0.07}} & 0.920$\pm$\scriptsize{0.05} & 0.851$\pm$\scriptsize{0.07} & 0.886$\pm$\scriptsize{0.07} & 0.868$\pm$\scriptsize{0.06} & 0.936$\pm$\scriptsize{0.07} & 0.833$\pm$\scriptsize{0.07} & \underline{\textbf{0.941$\pm$\scriptsize{0.06}}}  \\
glass6  &  0.901$\pm$\scriptsize{0.04} & 0.898$\pm$\scriptsize{0.04} & 0.903$\pm$\scriptsize{0.04} & \underline{\textbf{0.932$\pm$\scriptsize{0.03}}} & 0.903$\pm$\scriptsize{0.03} & 0.897$\pm$\scriptsize{0.03} & 0.915$\pm$\scriptsize{0.04} & 0.902$\pm$\scriptsize{0.03} & 0.928$\pm$\scriptsize{0.03} & 0.904$\pm$\scriptsize{0.04} & \underline{0.930$\pm$\scriptsize{0.03}}  \\
Heart D. &  \underline{0.800$\pm$\scriptsize{0.02}} & 0.798$\pm$\scriptsize{0.02} & 0.798$\pm$\scriptsize{0.02} & 0.795$\pm$\scriptsize{0.02} & \underline{\textbf{0.801$\pm$\scriptsize{0.02}}} & 0.784$\pm$\scriptsize{0.02} & 0.789$\pm$\scriptsize{0.02} & 0.789$\pm$\scriptsize{0.02} & 0.789$\pm$\scriptsize{0.02} & 0.797$\pm$\scriptsize{0.02} & 0.798$\pm$\scriptsize{0.02}  \\
N. Thyr.  &  \underline{0.957$\pm$\scriptsize{0.02}} & \underline{\textbf{0.967$\pm$\scriptsize{0.01}}} & 0.939$\pm$\scriptsize{0.03} & 0.943$\pm$\scriptsize{0.03} & 0.918$\pm$\scriptsize{0.03} & 0.901$\pm$\scriptsize{0.02} & 0.904$\pm$\scriptsize{0.03} & 0.902$\pm$\scriptsize{0.03} & 0.936$\pm$\scriptsize{0.03} & 0.926$\pm$\scriptsize{0.02} & 0.906$\pm$\scriptsize{0.03}  \\
yeast  &  \underline{\textbf{0.527$\pm$\scriptsize{0.02}}} & \underline{0.527$\pm$\scriptsize{0.02}} & 0.527$\pm$\scriptsize{0.02} & 0.512$\pm$\scriptsize{0.03} & 0.516$\pm$\scriptsize{0.03} & 0.424$\pm$\scriptsize{0.03} & 0.440$\pm$\scriptsize{0.03} & 0.504$\pm$\scriptsize{0.03} & 0.507$\pm$\scriptsize{0.03} & 0.498$\pm$\scriptsize{0.02} & 0.523$\pm$\scriptsize{0.03}  \\
yeast1  &  0.762$\pm$\scriptsize{0.04} & 0.755$\pm$\scriptsize{0.04} & 0.732$\pm$\scriptsize{0.04} & 0.741$\pm$\scriptsize{0.04} & 0.744$\pm$\scriptsize{0.04} & \underline{0.810$\pm$\scriptsize{0.04}} & \underline{\textbf{0.824$\pm$\scriptsize{0.03}}} & 0.737$\pm$\scriptsize{0.04} & 0.743$\pm$\scriptsize{0.04} & 0.795$\pm$\scriptsize{0.04} & 0.749$\pm$\scriptsize{0.05}  
\\\hline\hline
{\bf Rank} & \underline{4.14 (2-4-0)} & 4.96 (3-2-1) & 6.04 (0-1-2) & 6.61 (1-1-0) & 5.75 (1-1-1) & 7.36 (0-1-4) & 6.11 (2-0-1) & 8.86 (0-0-2) & 7.39 (0-0-2) & 5.29 (1-2-1) & \underline{\textbf{3.50 (4-2-0)}} \\\hline
\end{tabular}

\medskip

% -----------------------------------------------------------------------------------------------
% RF - BALANCED ACCURACY
% -----------------------------------------------------------------------------------------------
\begin{tabular}{|l||P{1.4cm}|P{1.4cm}|P{1.6cm}|P{1.4cm}|P{1.4cm}|P{1.5cm}|P{1.4cm}|P{1.4cm}|P{1.4cm}|P{1.4cm}|P{1.5cm}|}

\multicolumn{12}{c}{\bf Random Forest Classifier -- Balanced Accuracy -- Friedman test $p$-value: $1.811773\cdot 10^{-6}$}\\\hline
{\bf Dataset} & {\bf SMOTE} & {\bf ADASYN} & {\bf KM-SMOTE} & {\bf CGAN} & {\bf COPGAN} & {\bf CTAB-GAN} & {\bf CTGAN} & {\bf GCOP} & {\bf SB-GAN} & {\bf TVAE} & {\bf \ganname{}}  \\\hline

Anemia  &  \underline{\textbf{0.950$\pm$\scriptsize{0.02}}} & \underline{0.922$\pm$\scriptsize{0.02}} & 0.922$\pm$\scriptsize{0.02} & 0.888$\pm$\scriptsize{0.03} & 0.884$\pm$\scriptsize{0.02} & 0.877$\pm$\scriptsize{0.03} & 0.882$\pm$\scriptsize{0.03} & 0.904$\pm$\scriptsize{0.03} & 0.911$\pm$\scriptsize{0.03} & 0.887$\pm$\scriptsize{0.02} & 0.916$\pm$\scriptsize{0.02}  \\
Churn  &  \underline{0.744$\pm$\scriptsize{0.00}} & \underline{\textbf{0.747$\pm$\scriptsize{0.00}}} & 0.710$\pm$\scriptsize{0.01} & 0.702$\pm$\scriptsize{0.01} & 0.685$\pm$\scriptsize{0.01} & 0.705$\pm$\scriptsize{0.01} & 0.721$\pm$\scriptsize{0.01} & 0.707$\pm$\scriptsize{0.01} & 0.700$\pm$\scriptsize{0.01} & 0.719$\pm$\scriptsize{0.01} & 0.709$\pm$\scriptsize{0.01}  \\
Dry B.  &  \underline{\textbf{0.934$\pm$\scriptsize{0.00}}} & 0.933$\pm$\scriptsize{0.00} & 0.933$\pm$\scriptsize{0.00} & 0.932$\pm$\scriptsize{0.00} & 0.931$\pm$\scriptsize{0.00} & \underline{0.934$\pm$\scriptsize{0.00}} & 0.933$\pm$\scriptsize{0.00} & 0.930$\pm$\scriptsize{0.00} & 0.932$\pm$\scriptsize{0.00} & 0.929$\pm$\scriptsize{0.00} & 0.933$\pm$\scriptsize{0.00}  \\
ecoli1  &  \underline{\textbf{0.896$\pm$\scriptsize{0.02}}} & \underline{0.894$\pm$\scriptsize{0.02}} & 0.856$\pm$\scriptsize{0.03} & 0.869$\pm$\scriptsize{0.03} & 0.865$\pm$\scriptsize{0.03} & 0.856$\pm$\scriptsize{0.03} & 0.859$\pm$\scriptsize{0.03} & 0.864$\pm$\scriptsize{0.02} & 0.867$\pm$\scriptsize{0.03} & 0.873$\pm$\scriptsize{0.02} & 0.880$\pm$\scriptsize{0.03}  \\
ecoli2  &  0.904$\pm$\scriptsize{0.02} & 0.892$\pm$\scriptsize{0.02} & 0.883$\pm$\scriptsize{0.03} & 0.884$\pm$\scriptsize{0.03} & 0.869$\pm$\scriptsize{0.03} & 0.876$\pm$\scriptsize{0.04} & 0.877$\pm$\scriptsize{0.03} & 0.874$\pm$\scriptsize{0.03} & 0.899$\pm$\scriptsize{0.03} & \underline{0.908$\pm$\scriptsize{0.03}} & \underline{\textbf{0.910$\pm$\scriptsize{0.03}}}  \\
ecoli4  &  \underline{0.909$\pm$\scriptsize{0.04}} & 0.894$\pm$\scriptsize{0.04} & 0.843$\pm$\scriptsize{0.04} & 0.893$\pm$\scriptsize{0.04} & 0.872$\pm$\scriptsize{0.03} & 0.825$\pm$\scriptsize{0.05} & 0.878$\pm$\scriptsize{0.03} & 0.869$\pm$\scriptsize{0.03} & 0.900$\pm$\scriptsize{0.04} & 0.900$\pm$\scriptsize{0.06} & \underline{\textbf{0.928$\pm$\scriptsize{0.04}}}  \\
Fetal H. &  \underline{0.896$\pm$\scriptsize{0.01}} & \underline{\textbf{0.898$\pm$\scriptsize{0.01}}} & 0.868$\pm$\scriptsize{0.01} & 0.869$\pm$\scriptsize{0.01} & 0.869$\pm$\scriptsize{0.01} & 0.865$\pm$\scriptsize{0.01} & 0.866$\pm$\scriptsize{0.01} & 0.871$\pm$\scriptsize{0.01} & 0.867$\pm$\scriptsize{0.01} & 0.867$\pm$\scriptsize{0.01} & 0.871$\pm$\scriptsize{0.02}  \\
Flare-F  &  0.589$\pm$\scriptsize{0.03} & 0.591$\pm$\scriptsize{0.03} & 0.525$\pm$\scriptsize{0.02} & 0.554$\pm$\scriptsize{0.05} & 0.719$\pm$\scriptsize{0.04} & 0.531$\pm$\scriptsize{0.03} & 0.719$\pm$\scriptsize{0.04} & 0.558$\pm$\scriptsize{0.04} & 0.561$\pm$\scriptsize{0.06} & \underline{0.766$\pm$\scriptsize{0.04}} & \underline{\textbf{0.771$\pm$\scriptsize{0.03}}}  \\
glass5  &  0.871$\pm$\scriptsize{0.08} & 0.873$\pm$\scriptsize{0.06} & 0.757$\pm$\scriptsize{0.08} & 0.758$\pm$\scriptsize{0.10} & \underline{0.878$\pm$\scriptsize{0.08}} & 0.795$\pm$\scriptsize{0.08} & 0.873$\pm$\scriptsize{0.07} & 0.851$\pm$\scriptsize{0.09} & 0.786$\pm$\scriptsize{0.08} & 0.771$\pm$\scriptsize{0.07} & \underline{\textbf{0.933$\pm$\scriptsize{0.06}}}  \\
glass6  &  0.916$\pm$\scriptsize{0.03} & \underline{0.933$\pm$\scriptsize{0.03}} & 0.914$\pm$\scriptsize{0.03} & 0.895$\pm$\scriptsize{0.03} & 0.913$\pm$\scriptsize{0.04} & 0.925$\pm$\scriptsize{0.03} & 0.921$\pm$\scriptsize{0.03} & 0.918$\pm$\scriptsize{0.03} & 0.900$\pm$\scriptsize{0.03} & 0.917$\pm$\scriptsize{0.04} & \underline{\textbf{0.937$\pm$\scriptsize{0.03}}}  \\
Heart D. &  \underline{0.817$\pm$\scriptsize{0.02}} & 0.812$\pm$\scriptsize{0.02} & 0.815$\pm$\scriptsize{0.03} & 0.814$\pm$\scriptsize{0.02} & 0.806$\pm$\scriptsize{0.02} & 0.812$\pm$\scriptsize{0.02} & 0.810$\pm$\scriptsize{0.02} & 0.813$\pm$\scriptsize{0.02} & 0.817$\pm$\scriptsize{0.02} & 0.813$\pm$\scriptsize{0.02} & \underline{\textbf{0.827$\pm$\scriptsize{0.02}}}  \\
N. Thyr.  &  0.932$\pm$\scriptsize{0.03} & 0.915$\pm$\scriptsize{0.03} & 0.930$\pm$\scriptsize{0.03} & 0.926$\pm$\scriptsize{0.02} & 0.917$\pm$\scriptsize{0.03} & 0.909$\pm$\scriptsize{0.02} & \underline{0.932$\pm$\scriptsize{0.02}} & 0.905$\pm$\scriptsize{0.03} & 0.923$\pm$\scriptsize{0.02} & \underline{\textbf{0.942$\pm$\scriptsize{0.01}}} & 0.931$\pm$\scriptsize{0.02}  \\
yeast  &  0.528$\pm$\scriptsize{0.03} & 0.528$\pm$\scriptsize{0.03} & 0.528$\pm$\scriptsize{0.03} & \underline{0.563$\pm$\scriptsize{0.02}} & 0.548$\pm$\scriptsize{0.03} & 0.551$\pm$\scriptsize{0.03} & 0.537$\pm$\scriptsize{0.03} & 0.556$\pm$\scriptsize{0.03} & 0.561$\pm$\scriptsize{0.03} & 0.561$\pm$\scriptsize{0.02} & \underline{\textbf{0.567$\pm$\scriptsize{0.04}}}  \\
yeast1  &  \underline{0.768$\pm$\scriptsize{0.05}} & \underline{\textbf{0.781$\pm$\scriptsize{0.05}}} & 0.677$\pm$\scriptsize{0.04} & 0.684$\pm$\scriptsize{0.04} & 0.667$\pm$\scriptsize{0.04} & 0.670$\pm$\scriptsize{0.04} & 0.702$\pm$\scriptsize{0.04} & 0.670$\pm$\scriptsize{0.04} & 0.670$\pm$\scriptsize{0.04} & 0.719$\pm$\scriptsize{0.05} & 0.679$\pm$\scriptsize{0.04}  \\
\hline\hline

{\bf Rank} & \underline{3.36 (3-5-1)} & 4.46 (3-3-0) & 7.39 (0-0-3) & 6.71 (0-1-1) & 7.96 (0-1-4) & 8.14 (0-1-3) & 6.11 (0-1-0) & 7.50 (0-0-1) & 6.71 (0-0-0) & 5.07 (1-2-1) & \underline{\textbf{2.57 (7-0-0)}} \\\hline
\end{tabular}

\end{center}
\end{table*}

The results indicate that \ganname{} achieved the best overall average ranking among the evaluated
methods across the considered oversampling tasks, classifiers, and evaluation measures. For 
example, XGBoost and RF achieved their best $B_{ac}$ values in 7 out of the \numdatasets{} datasets
that have been balanced by \ganname{}. On average, our model was ranked 2.4th and 2.6th in the
$B_{ac}$ measurements of XGBoost and RF respectively. Regarding $F1$, \ganname{} again achieved the
strongest overall behavior by achieving top performance in 6 datasets and second best performance 
in two cases.

SMOTE and ADASYN were particularly strong in this task, outperforming the other generative models
in most cases. However, these methods are specifically designed for class imbalance and not full
data generation. Compared to the generative models that aim for realism and privacy preservation, 
these data mining methods solve a narrower problem.

The most competitive generative models were TVAE and ctGAN. The former was ranked 5.1st and 5.7th
in the $B_{ac}$ measurements of RF and XGBoost. Compared to our model, it achieved the best
performance only in one dataset in each test, apart from $B_{ac}$ of XGBoost where it was always
ranked below the first place. ctGAN was the second strongest generative model in terms of $B_{ac}$
for MLP and RF. However, \ganname{} achieved better overall results than both models in the
reported classification experiments.

% -----------------------------------------------------------------------------------------------
% -----------------------------------------------------------------------------------------------
% FIDELITY EXPERIMENT - CLASSIFICATION PERFORMANCE
% -----------------------------------------------------------------------------------------------
% -----------------------------------------------------------------------------------------------

% -----------------------------------------------------------------------------------------------
% MLP FIDELITY
% -----------------------------------------------------------------------------------------------
\begin{table*}
\begin{center}
\caption{Results for the data fidelity experiment. The tables illustrate the percent difference
(\%) between the performance of various classifiers in real and synthetic datasets created by the
examined generative models. The top table concerns the Balanced Accuracy and $F1$-score of XGBoost.
The bottom table concerns the Balanced Accuracy of MLP and Random Forest classifiers.}
\label{tab:fid}
\setlength\tabcolsep{3.0pt}

% -----------------------------------------------------------------------------------------------
% XGBOOST FIDELITY
% -----------------------------------------------------------------------------------------------
\begin{tabular}{|l||P{0.85cm}P{0.80cm}P{0.85cm}P{0.80cm}P{0.80cm}P{0.80cm}P{0.80cm}P{0.80cm}||P{0.85cm}P{0.80cm}P{0.85cm}P{0.80cm}P{0.80cm}P{0.80cm}P{0.80cm}P{0.80cm}|} \hline
\multirow{3}{*}{\bf Dataset} & \multicolumn{8}{c||}{\bf XGBoost -- Bal. Acc. -- Friedman test $p$-value: $5.98878\cdot 10^{-5}$} & \multicolumn{8}{c|}{\bf XGBoost -- $F1$ score -- Friedman Test $p$-value: $5.3860\cdot 10^{-11}$} \\\cline{2-9}\cline{10-17}
 & \multirow{2}{*}{\bf CGAN} & {\bf COP GAN} & {\bf CTAB GAN+} & {\bf CT GAN} & \multirow{2}{*}{\bf GCOP} & {\bf SB GAN} & \multirow{2}{*}{\bf TVAE} & {\bf ctd GAN} & \multirow{2}{*}{\bf CGAN} & {\bf COP GAN} & {\bf CTAB GAN+} & {\bf  CT GAN} & \multirow{2}{*}{\bf GCOP} & {\bf SB GAN} &  \multirow{2}{*}{\bf TVAE} & {\bf ctd GAN} \\\hline

Anemia  &  -89.1 & -54.3 & -63.4 & \underline{-51.1} & -84.0 & -89.1 & \underline{\textbf{-40.1}} & -51.8  &  -77.5 & -29.3 & -45.8 & \underline{-26.0} & -72.1 & -77.8 & \underline{\textbf{-22.0}} & -39.5  \\
Churn &  -30.6 & \underline{\textbf{-8.6}} & 14.6 & \underline{-10.0} & -29.3 & -30.8 & 29.2 & 14.0 &  -91.4 & \underline{-29.5} & 33.8 & \underline{\textbf{-23.2}} & -81.6 & -95.0 & 55.2 & 63.5  \\
Dry Bean  &  -84.8 & -7.0 & -11.7 & -5.9 & -70.2 & -84.6 & \underline{\textbf{-1.5}} & \underline{5.0}  &  -77.5 & -7.9 & -11.7 & -6.6 & -59.7 & -76.4 & \underline{\textbf{-1.6}} & \underline{5.0}  \\
ecoli1  &  -44.1 & -1.1 & -14.7 & \underline{0.5} & -23.7 & -41.7 & -4.0 & \underline{\textbf{-0.2}} &  -80.0 & \underline{3.4} & -24.9 & \underline{\textbf{0.9}} & -40.9 & -90.5 & -8.8 & -5.4  \\
ecoli2  &  -42.0 & -20.2 & -22.2 & -10.6 & -37.3 & -42.1 & \underline{-8.4} & \underline{\textbf{-1.9}} &  -86.3 & -35.5 & -38.7 & -20.2 & -75.2 & -88.9 & \underline{-19.7} & \underline{\textbf{-15.3}}  \\
ecoli4  &  -40.0 & \underline{-8.0} & -31.2 & \underline{\textbf{-7.7}} & -39.2 & -39.7 & -16.5 & 7.7 &  -100.0 & -20.9 & -76.0 & \underline{\textbf{-14.2}} & -95.9 & -100.0 & \underline{-19.6} & -29.7  \\
Fetal Heal. &  -63.7 & \underline{\textbf{-11.4}} & -30.5 & \underline{-14.2} & -38.2 & -63.4 & -14.4 & -17.3 &  -16.0 & \underline{\textbf{-0.8}} & -8.7 & \underline{-1.5} & -8.7 & -13.5 & -1.8 & -3.0  \\
Flare-F  &  \underline{-12.4} & 32.9 & -12.6 & 49.0 & -13.1 & -12.6 & 68.0 & \underline{\textbf{6.1}}   &  \underline{-100.0} & 168.4 & -100.0 & 258.5 & -100.0 & -100.0 & 338.2 & \underline{\textbf{14.6}}  \\
glass5  &  -29.6 & -28.3 & -30.0 & -28.3 & -29.6 & -29.8 & \underline{\textbf{-2.0}} & \underline{-21.2} &  -100.0 & -100.0 & -100.0 & -93.9 & -100.0 & -100.0 & \underline{\textbf{-7.1}} & \underline{-75.3}  \\
glass6  &  -44.6 & -32.5 & -25.3 & -32.3 & -33.9 & -44.8 & \underline{\textbf{-0.0}} & \underline{-3.5}  &  -100.0 & -72.0 & -48.2 & -67.1 & -67.0 & -97.7 & \underline{\textbf{-6.8}} & \underline{-43.7}  \\
Heart Dis. &  -37.8 & -14.8 & \underline{\textbf{-6.2}} & -18.7 & -22.9 & -36.6 & 13.1 & \underline{-11.9}  &  -33.3 & -11.2 & \underline{\textbf{-3.9}} & -17.1 & -21.5 & -32.0 & 12.6 & \underline{-6.9}  \\
N. Thyroid  &  -65.0 & -50.8 & -46.8 & -53.4 & -44.5 & -64.1 & \underline{-22.0} & \underline{\textbf{-7.2}} &  -33.7 & -28.1 & -24.5 & -29.0 & -22.7 & -34.8 & \underline{-13.6} & \underline{\textbf{-4.8}}  \\
yeast  &  -82.6 & -43.4 & -52.7 & -45.5 & -77.8 & -83.0 & \underline{-2.0} & \underline{\textbf{-0.9}}  &  -52.1 & -16.8 & -21.2 & \underline{-10.8} & -45.2 & -44.5 & 15.6 & \underline{\textbf{1.8}}  \\
yeast1  &  -31.7 & \underline{\textbf{-2.7}} & -28.0 & 9.0 & -31.8 & -31.7 & 8.9 & \underline{3.1}  &  -100.0 & \underline{-32.0} & -88.9 & 37.0 & -100.0 & -100.0 & \underline{\textbf{26.9}} & -54.8  \\

\hline\hline
\multirow{2}{*}{\textbf{Rank}}  &  6.96 (0-1-6) & 3.39 (3-1-0) & 4.54 (1-0-1) & 3.50 (1-4-0) & 5.82 (0-0-1) & 7.04 (0-0-5) & \underline{2.86} \underline{(4-3-1)} & \underline{\textbf{1.89}} \underline{\textbf{(5-5-0)}} & 7.07 (0-1-7) & 3.64 (1-3-0) & 4.35 (1-0-0) & 3.14 (3-3-0) & 5.57 (0-0-0) & 7.00 (0-0-6) & \underline{2.64} \underline{(5-3-1)} & \underline{\textbf{2.57}} \underline{\textbf{(4-4-0)}}  \\\hline
\end{tabular}

\medskip

% -----------------------------------------------------------------------------------------------
% MLP BACC / RF BACC FIDELITY
% -----------------------------------------------------------------------------------------------
\begin{tabular}{|l||P{0.85cm}P{0.80cm}P{0.85cm}P{0.80cm}P{0.80cm}P{0.80cm}P{0.80cm}P{0.80cm}||P{0.85cm}P{0.80cm}P{0.85cm}P{0.80cm}P{0.80cm}P{0.80cm}P{0.80cm}P{0.80cm}|} \hline
\multirow{3}{*}{\bf Dataset} & \multicolumn{8}{c||}{\bf MLP -- Bal. Accuracy -- Friedman test $p$-value: $1.29716\cdot 10^{-8}$} & \multicolumn{8}{c|}{\bf RF -- Bal. Accuracy -- Friedman Test $p$-value: $1.46346\cdot 10^{-5}$} \\\cline{2-9}\cline{10-17}
 & \multirow{2}{*}{\bf CGAN} & {\bf COP GAN} & {\bf CTAB GAN+} & {\bf CT GAN} & \multirow{2}{*}{\bf GCOP} & {\bf SB GAN} & \multirow{2}{*}{\bf TVAE} & {\bf ctd GAN} & \multirow{2}{*}{\bf CGAN} & {\bf COP GAN} & {\bf CTAB GAN+} & {\bf  CT GAN} & \multirow{2}{*}{\bf GCOP} & {\bf SB GAN} &  \multirow{2}{*}{\bf TVAE} & {\bf ctd GAN} \\\hline

Anemia  &  -80.4 & -49.0 & -53.7 & \underline{-41.7} & -72.0 & -80.4 & \underline{\textbf{-33.3}} & -42.8 &  -88.1 & -57.5 & -64.6 & -53.5 & -84.3 & -88.4 & \underline{\textbf{-43.2}} & \underline{-53.1} \\
Churn &  -1.8 & 3.3 & 6.7 & \underline{0.8} & -2.2 & -2.0 & 1.8 & \underline{\textbf{0.2}} &  -29.7 & \underline{\textbf{-9.4}} & 12.8 & \underline{-11.0} & -29.0 & -29.6 & 31.3 & 12.7 \\
Dry Bean  &  -74.6 & -39.1 & -33.3 & -41.5 & -69.2 & -75.0 & \underline{-31.3} & \underline{\textbf{-14.8}} &  -84.7 & -8.3 & -11.8 & -7.4 & -70.5 & -84.8 & \underline{\textbf{-0.90}} & \underline{4.3} \\
ecoli1  &  -41.2 & \underline{\textbf{0.4}} & -12.6 & 2.9 & -19.3 & -41.2 & \underline{-0.5} & 1.5 &  -40.9 & \underline{-0.8} & -12.0 & 1.9 & -20.4 & -40.9 & \underline{\textbf{-0.7}} & 3.0  \\
ecoli2  &  -44.4 & -21.9 & -22.1 & -13.8 & -41.8 & -44.4 & \underline{-7.0} & \underline{\textbf{0.4}} &  -42.8 & -24.0 & -28.5 & -16.6 & -39.4 & -42.9 & \underline{-12.0} & \underline{\textbf{-3.20}}  \\
ecoli4  &  -44.5 & -18.9 & -40.0 & \underline{-10.0} & -44.5 & -44.5 & -18.5 & \underline{\textbf{9.1}} &  -40.1 & -21.2 & -38.7 & \underline{-20.3} & -40.7 & -40.7 & -29.2 & \underline{\textbf{-0.20}}  \\
Fetal Heal. &  -56.5 & -9.4 & -27.0 & \underline{-9.2} & -26.3 & -56.1 & \underline{\textbf{-7.6}} & -16.4  &  -61.9 & \underline{\textbf{-12.70}} & -33.4 & -17.5 & -38.5 & -61.9 & \underline{-15.8} & -22.2 \\
Flare-F  &  -11.7 & 36.6 & \underline{-10.2} & 49.8 & -12.2 & -11.7 & 67.3 & \underline{\textbf{5.7}}  &  \underline{-4.4} & 43.3 & \underline{\textbf{-4.20}} & 61.7 & -5.2 & -4.4 & 82.4 & 13.2  \\
glass5  &  \underline{-16.7} & -16.8 & -16.8 & -17.0 & -16.7 & -16.7 & \underline{\textbf{-0.2}} & -16.8  &  \underline{-35.2} & -35.4 & \underline{\textbf{-33.8}} & -35.2 & -35.2 & -35.2 & -35.2 & -35.2 \\
glass6  &  -16.8 & -14.6 & -6.0 & -15.6 & \underline{-0.8} & -16.8 & \underline{\textbf{0.2}} & -9.5 &  -43.5 & -36.1 & -27.9 & -36.9 & -32.5 & -43.5 & \underline{\textbf{4.00}} & \underline{-9.0} \\
Heart Dis. &  -33.2 & -14.5 & \underline{-9.0} & -15.7 & -21.5 & -33.5 & \underline{\textbf{7.5}} & -13.2 &  -39.7 & \underline{-13.4} & \underline{\textbf{-7.20}} & -16.0 & -23.7 & -37.8 & 13.9 & -14.7 \\
N. Thyroid  &  -65.4 & -57.8 & -52.7 & -63.6 & -54.5 & -65.4 & \underline{-30.1} & \underline{\textbf{-9.0}} &  -65.0 & -50.0 & -47.8 & -57.8 & -48.5 & -67.1 & \underline{-22.6} & \underline{\textbf{-6.40}}  \\
yeast  &  -81.8 & -38.9 & -47.2 & -45.1 & -77.6 & -82.9 & \underline{-11.6} & \underline{\textbf{-2.6}} &  -81.8 & -42.7 & -49.9 & -44.2 & -76.9 & -82.2 & \underline{-8.7} & \underline{\textbf{-1.50}}  \\
yeast1  &  -26.0 & \underline{\textbf{6.5}} & -26.0 & 17.5 & -26.0 & -26.0 & 14.1 & \underline{6.7} &  -25.8 & \underline{-1.5} & -24.2 & \underline{\textbf{0.2}} & -25.8 & -25.8 & 8.9 & -1.9   \\

\hline\hline
\multirow{2}{*}{\textbf{Rank}} &  6.43 (0-1-5) & 4.07 (2-0-0) & 4.61 (0-2-1) & 4.29 (0-4-1) & 5.32 (0-1-1) & 6.68 (0-0-6) & \underline{2.32} \underline{(5-5-0)} & \underline{\textbf{2.28}} \underline{\textbf{(7-1-0)}} &  6.57 (0-2-4) & 3.50 (2-3-0) & 3.79 (3-0-0) & 3.71 (1-2-0) & 5.54 (0-0-2) & 6.89 (0-0-6) & \underline{3.07} \underline{(4-4-2)} & \underline{\textbf{2.93}} \underline{\textbf{(4-3-0)}} \\\hline
\end{tabular}
\end{center}
\end{table*}

\subsection{Synthetic Data Fidelity}
\label{ssec:exp-detect}

The data fidelity experiment evaluates the ability of a model to synthesize realistic data. We 
applied two techniques for this purpose: i) similarity of classification performance (also called 
ML utility), and ii) comparison of column-wise correlations. The first compares the performance 
of a classifier trained and tested with real data, and the performance of the same model trained 
and tested on synthetic data. The second test computes the column correlations of the real and
synthetic data and examines their difference.

The first technique comprises a five steps shown in Fig.~\ref{fig:exp-fid}: Initially, a classifier
$\mathcal{P}$ is trained and tested on the original dataset $\mathbf{X}$. Then, a generative model
$\mathcal{M}$ is trained on $\mathbf{X}$ and synthesizes a new dataset $\mathbf{\hat X}$ with the
same class distribution as $\mathbf{X}$. Next, a classifier $\mathcal{P'}$ (with identical 
properties as $\mathcal{P}$) is trained and tested on the synthetic dataset. In the final stage,
the performances of $\mathcal{P}$ and $\mathcal{P'}$ are compared: the smaller their discrepancy,
the more realistic the synthetic data is.

The results are presented in Table~\ref{tab:fid}. These values represent the percentage differences
between $F1$ and $F1'$, and $B_{ac}$ and $B'_{ac}$, when the predictors $\mathcal{P}$ and
$\mathcal{P}'$ are XGBoost (top), MLP, or RF classifiers (bottom). Similarly to the previous case,
\ganname{} exhibited strong performance and generated highly realistic synthetic samples. Compared
to the other evaluated models, the classification performance differences between the original and
the synthetic datasets of \ganname{} were the smallest. This applies to both the $F1$ and $B_{ac}$
measures.

The strongest opponents were TVAE, ctGAN and Copula GAN; they all produced data of high fidelity,
but inferior to that of \ganname. On the other hand, the Conditional GAN and SB-GAN generated
rather unrealistic data. In some cases, the performance difference between the real and synthetic
datasets approached $-100\%$. We shall keep this observation for later, when we compare the models
in terms of privacy preservation.

\begin{table*}
\begin{center}
\caption{Comparison of model performance in terms of column correlation distances. Smaller
distances indicate a greater capability to preserve feature correlations. The heatmaps illustrate
the correlation structures in the New-Thyroid and Yeast datasets.}
\label{tab:ccorr}
\setlength\tabcolsep{3.0pt}
% -----------------------------------------------------------------------------------------------
% DIFFERENCE IN COLUMN CORRELATIONS
% -----------------------------------------------------------------------------------------------

\begin{tabular}{|l||P{1.7cm}|P{1.7cm}|P{1.7cm}|P{1.7cm}|P{1.7cm}|P{1.7cm}|P{1.7cm}|P{1.7cm}|}
\multicolumn{9}{c}{\bf Mean column correlation difference (MCCD) between real and synthetic datasets -- Friedman test $p$-value: $9.508103\cdot 10^{-4}$}\\\hline
{\bf Dataset} & {\bf CGAN} & {\bf COPGAN} & {\bf CTGAN} & {\bf CTAB-GAN+} & {\bf GCOP} & {\bf SB-GAN} & {\bf TVAE} & {\bf \ganname{}}  \\\hline

Anemia  &  0.2131$\pm$\scriptsize{0.20} & \underline{0.1239$\pm$\scriptsize{0.01}} & \underline{\textbf{0.1192$\pm$\scriptsize{0.02}}} & 0.1378$\pm$\scriptsize{0.01} & 0.1318$\pm$\scriptsize{0.00} & 0.2079$\pm$\scriptsize{0.25} & 0.1765$\pm$\scriptsize{0.03} & 0.2102$\pm$\scriptsize{0.10}  \\
Churn Model.  &  0.0639$\pm$\scriptsize{0.06} & 0.0456$\pm$\scriptsize{0.01} & \underline{0.0424$\pm$\scriptsize{0.00}} & 0.5916$\pm$\scriptsize{0.00} & \underline{\textbf{0.0220$\pm$\scriptsize{0.00}}} & 0.0619$\pm$\scriptsize{0.05} & 0.0837$\pm$\scriptsize{0.03} & 0.1482$\pm$\scriptsize{0.03}  \\
Dry Bean  &  0.2461$\pm$\scriptsize{0.25} & 0.1470$\pm$\scriptsize{0.02} & 0.1247$\pm$\scriptsize{0.01} & \underline{\textbf{0.0455$\pm$\scriptsize{0.02}}} & 0.0798$\pm$\scriptsize{0.01} & 0.2481$\pm$\scriptsize{0.22} & \underline{0.0704$\pm$\scriptsize{0.02}} & 0.0706$\pm$\scriptsize{0.02}  \\
ecoli1  &  0.1064$\pm$\scriptsize{0.04} & 0.1439$\pm$\scriptsize{0.03} & 0.1392$\pm$\scriptsize{0.02} & 0.0822$\pm$\scriptsize{0.02} & \underline{\textbf{0.0372$\pm$\scriptsize{0.01}}} & 0.1205$\pm$\scriptsize{0.02} & 0.0802$\pm$\scriptsize{0.02} & \underline{0.0731$\pm$\scriptsize{0.02}}  \\
ecoli2  &  0.1261$\pm$\scriptsize{0.05} & 0.1699$\pm$\scriptsize{0.03} & 0.1574$\pm$\scriptsize{0.02} & 0.0753$\pm$\scriptsize{0.01} & \underline{\textbf{0.0417$\pm$\scriptsize{0.00}}} & 0.1190$\pm$\scriptsize{0.02} & 0.0802$\pm$\scriptsize{0.02} & \underline{0.0683$\pm$\scriptsize{0.01}}  \\
ecoli4  &  0.1116$\pm$\scriptsize{0.06} & 0.1778$\pm$\scriptsize{0.03} & 0.1618$\pm$\scriptsize{0.02} & 0.0968$\pm$\scriptsize{0.06} & \underline{\textbf{0.0385$\pm$\scriptsize{0.00}}} & 0.1232$\pm$\scriptsize{0.06} & 0.0891$\pm$\scriptsize{0.01} & \underline{0.0866$\pm$\scriptsize{0.03}}  \\
Fetal Health  &  0.1865$\pm$\scriptsize{0.07} & 0.1563$\pm$\scriptsize{0.01} & 0.1205$\pm$\scriptsize{0.01} & 0.1354$\pm$\scriptsize{0.02} & 0.1260$\pm$\scriptsize{0.00} & 0.1859$\pm$\scriptsize{0.07} & \underline{\textbf{0.0553$\pm$\scriptsize{0.01}}} & \underline{0.1124$\pm$\scriptsize{0.01}}  \\
Flare-F  &  0.1579$\pm$\scriptsize{0.06} & 0.1462$\pm$\scriptsize{0.02} & \underline{0.1252$\pm$\scriptsize{0.10}} & 0.7141$\pm$\scriptsize{0.00} & 0.1674$\pm$\scriptsize{0.01} & 0.1787$\pm$\scriptsize{0.07} & 0.1453$\pm$\scriptsize{0.05} & \underline{\textbf{0.0881$\pm$\scriptsize{0.03}}}  \\
glass5  &  0.2249$\pm$\scriptsize{0.03} & 0.1934$\pm$\scriptsize{0.02} & 0.2055$\pm$\scriptsize{0.02} & 0.1685$\pm$\scriptsize{0.02} & \underline{\textbf{0.1084$\pm$\scriptsize{0.01}}} & 0.2145$\pm$\scriptsize{0.02} & 0.1720$\pm$\scriptsize{0.01} & \underline{0.1413$\pm$\scriptsize{0.02}}  \\
glass6  &  0.2272$\pm$\scriptsize{0.03} & 0.2007$\pm$\scriptsize{0.03} & 0.2050$\pm$\scriptsize{0.01} & 0.1560$\pm$\scriptsize{0.04} & \underline{\textbf{0.0914$\pm$\scriptsize{0.02}}} & 0.2205$\pm$\scriptsize{0.05} & 0.1576$\pm$\scriptsize{0.03} & \underline{0.1525$\pm$\scriptsize{0.06}}  \\
Heart Disease  &  0.1438$\pm$\scriptsize{0.02} & 0.1128$\pm$\scriptsize{0.01} & 0.1160$\pm$\scriptsize{0.00} & 0.5748$\pm$\scriptsize{0.00} & \underline{\textbf{0.0694$\pm$\scriptsize{0.01}}} & 0.1296$\pm$\scriptsize{0.02} & 0.1207$\pm$\scriptsize{0.00} & \underline{0.1063$\pm$\scriptsize{0.02}}  \\
New Thyroid  &  0.2554$\pm$\scriptsize{0.08} & 0.2987$\pm$\scriptsize{0.05} & 0.2939$\pm$\scriptsize{0.03} & 0.0963$\pm$\scriptsize{0.04} & \underline{0.0755$\pm$\scriptsize{0.02}} & 0.2568$\pm$\scriptsize{0.06} & 0.1252$\pm$\scriptsize{0.04} & \underline{\textbf{0.0369$\pm$\scriptsize{0.01}}}  \\
yeast  &  0.1313$\pm$\scriptsize{0.05} & 0.0410$\pm$\scriptsize{0.00} & 0.0529$\pm$\scriptsize{0.11} & \underline{0.0338$\pm$\scriptsize{0.04}} & \underline{\textbf{0.0164$\pm$\scriptsize{0.02}}} & 0.0941$\pm$\scriptsize{0.25} & 0.0524$\pm$\scriptsize{0.03} & 0.0651$\pm$\scriptsize{0.03}  \\
yeast1  &  0.1482$\pm$\scriptsize{0.07} & 0.0521$\pm$\scriptsize{0.01} & 0.0671$\pm$\scriptsize{0.01} & \underline{0.0283$\pm$\scriptsize{0.00}} & \underline{\textbf{0.0190$\pm$\scriptsize{0.00}}} & 0.1337$\pm$\scriptsize{0.02} & 0.0759$\pm$\scriptsize{0.01} & 0.0595$\pm$\scriptsize{0.02}  \\

\hline\hline
{\bf Mean rank}  &  6.57 (0-0-6) & 5.14 (0-1-4) & 4.79 (1-2-0) & 4.14 (1-2-3) & \underline{\textbf{2.00}}  \underline{\textbf{(9-1-0)}}& 6.36 (0-0-1) & 3.86 (1-1-0) & \underline{3.14} \underline{(2-7-0)}  \\\hline
\end{tabular}

\bigskip

\center
\includegraphics[scale=0.232]{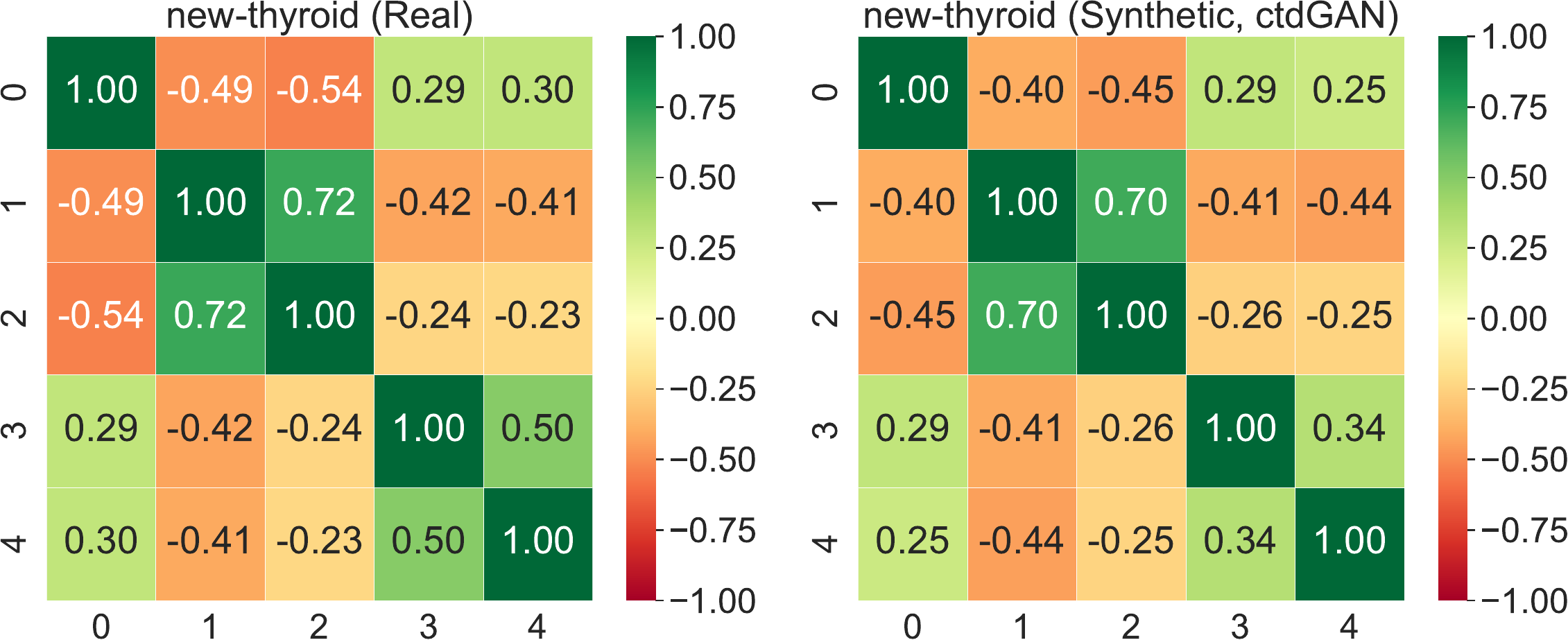}\hspace{5pt}
\includegraphics[scale=0.232]{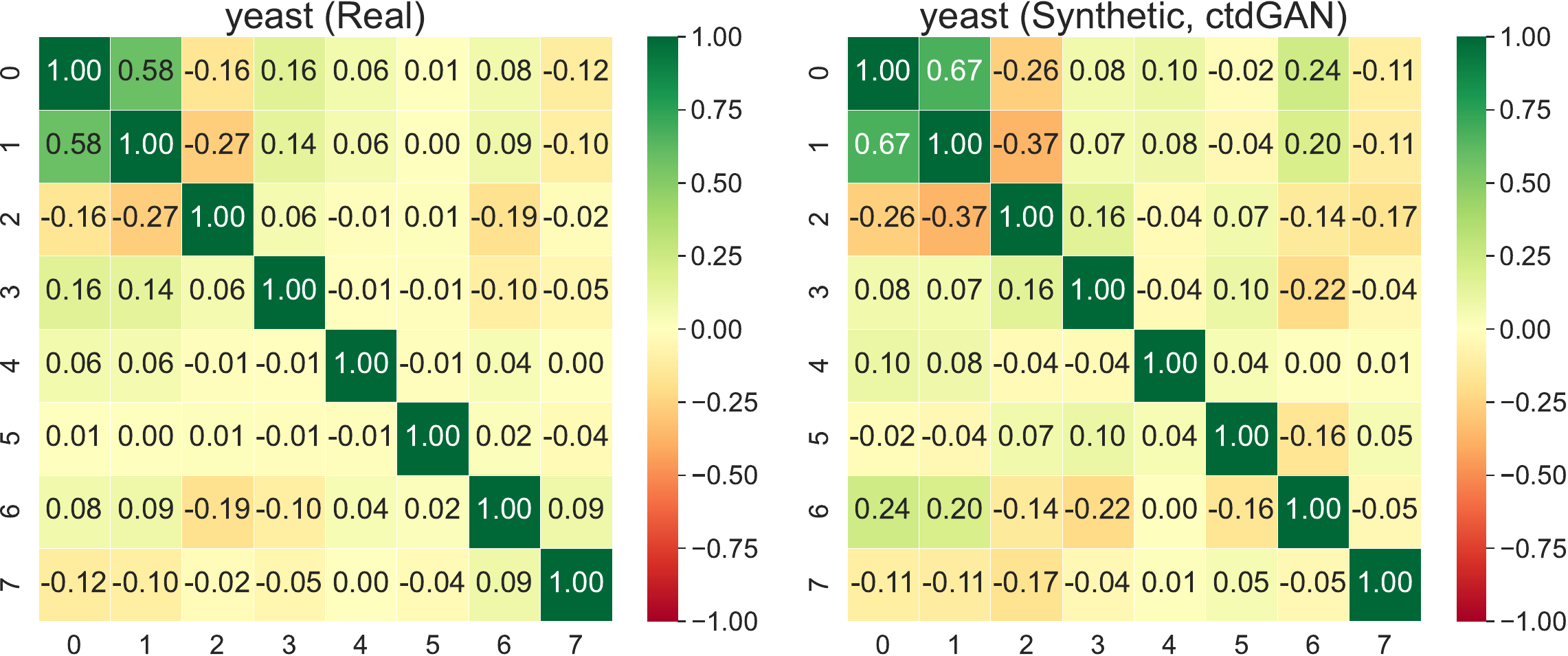}
\end{center}
\end{table*}

The second experiment on data fidelity was carried out by comparing the column correlations of a 
real dataset $\mathbf{X}$ with the column correlations of a synthetic dataset $\mathbf{\hat X}$ 
that was generated by a model $\mathcal{M}$ trained on $\mathbf{X}$. Similarly to the first test,
$\mathbf{X}$ and $\mathbf{\hat X}$ share a common class distribution.

Since the columns can be of mixed types, we employed a composite measure that quantifies the 
correlations by using: i) the Pearson coefficient for the comparison of numeric columns, ii)
Cramér's V for the discrete columns, and iii) the Correlation Ratio $\eta^2$ for comparing numeric
with categorical columns. This measure was used to create the correlation matrices $M^{\mathbf{X}}$
and $M^{\mathbf{\hat X}}$ for the real and synthetic datasets respectively. We now define Mixed
Column Correlation Difference (MCCD) as the mean of the absolute values of the matrix
$\Delta M=M^{\mathbf{X}}-M^{\mathbf{\hat X}}$ and we use it to quantify the difference between the 
column-wise correlations between $\mathbf{X}$ and $\mathbf{\hat X}$.

The values of MCCD are presented in Table~\ref{tab:ccorr}. The Gaussian Copula method achieved the
smallest discrepancies in column correlations between $\mathbf{X}$ and $\mathbf{\hat X}$, followed
by \ganname{}, TVAE and ctGAN. The heatmaps below this table indicatively depict how the
column-wise correlations in the real dataset compare with those in the respective synthetic 
dataset.

\subsection{Memorization and Privacy Preservation}
\label{ssec:exp-mpt}

Data memorization is an undesired property exhibited by various generative models. A model that
memorizes its training examples usually produces high fidelity synthetic samples not because it
effectively learned the underlying data distribution, but because it simply copies, or mimics its
input. In such cases, the model exhibits poor generalization and may even leak private records 
during sample generation.

% -----------------------------------------------------------------------------------------------
% PRIVACY PRESERVATION/MEMORIZATION
% -----------------------------------------------------------------------------------------------
\begin{table*}
\begin{center}
\caption{Memorization/Privacy tests with Distance to Nearest Neighbor (DNN, left) and Nearest
Neighbor Distance Ratio (NNDR, right).}
\label{tab:prv}
\setlength\tabcolsep{3.0pt}

% -----------------------------------------------------------------------------------------------
% NN1 DISTANCES AND NNDR RATIO
% -----------------------------------------------------------------------------------------------
\begin{tabular}{|l||P{0.85cm}P{0.80cm}P{0.85cm}P{0.80cm}P{0.80cm}P{0.80cm}P{0.80cm}P{0.80cm}||P{0.85cm}P{0.80cm}P{0.85cm}P{0.80cm}P{0.80cm}P{0.80cm}P{0.80cm}P{0.80cm}|} \hline
\multirow{3}{*}{\bf Dataset} & \multicolumn{8}{c||}{\bf NN Distance (DNN) -- Friedman test $p$-value: $3.856760\cdot 10^{-5}$} & \multicolumn{8}{c|}{\bf NN Distance Ratio (NNDR) -- Friedman Test: $1.087024\cdot 10^{-7}$} \\\cline{2-9}\cline{10-17}
 & \multirow{2}{*}{\bf CGAN} & {\bf COP GAN} & {\bf CT GAN} & {\bf CTAB GAN+} & \multirow{2}{*}{\bf GCOP} & {\bf SB GAN} & \multirow{2}{*}{\bf TVAE} & {\bf ctd GAN} & \multirow{2}{*}{\bf CGAN} & {\bf COP GAN} & {\bf CT GAN} & {\bf CTAB GAN+} & \multirow{2}{*}{\bf GCOP} & {\bf SB GAN} &  \multirow{2}{*}{\bf TVAE} & {\bf ctd GAN} \\\hline

Anemia  &  \underline{0.0238} & 0.0056 & 0.0059 & 0.0065 & 0.0102 & \underline{\textbf{0.0375}} & 0.0043 & 0.0197 &  0.9028 & 0.8714 & 0.8834 & 0.8710 & 0.9006 & \underline{\textbf{0.9305}} & 0.8608 & \underline{0.9223}  \\
Churn &  0.0644 & 0.0369 & 0.0285 & \underline{\textbf{0.4152}} & 0.0500 & \underline{0.1837} & 0.0277 & 0.0439  &  0.8614 & 0.8103 & 0.8032 & \underline{\textbf{0.9912}} & 0.8407 & \underline{0.9501} & 0.7984 & 0.8563 \\
Dry Bean  &  \underline{\textbf{0.0977}} & 0.0582 & 0.0298 & 0.0176 & 0.0427 & \underline{0.0874} & 0.0179 & 0.0197 &  0.9196 & \underline{\textbf{0.9659}} & 0.9321 & 0.9167 & \underline{0.9581} & 0.9394 & 0.9229 & 0.9404 \\
ecoli1  &  0.0444 & \underline{\textbf{0.2319}} & 0.0436 & 0.0321 & \underline{0.2014} & 0.0586 & 0.0294 & 0.0346 &  \underline{0.8681} & 0.8653 & 0.8453 & 0.8237 & 0.7381 & \underline{\textbf{0.8859}} & 0.8285 & 0.8283 \\
ecoli2  &  0.0462 & \underline{\textbf{0.2383}} & 0.0434 & 0.0303 & \underline{0.2035} & 0.0592 & 0.0290 & 0.0355 &  \underline{0.8857} & 0.8814 & 0.8408 & 0.8331 & 0.7427 & \underline{\textbf{0.9032}} & 0.8336 & 0.8278 \\
ecoli4  &  0.0516 & \underline{\textbf{0.2434}} & 0.0466 & 0.0335 & \underline{0.2046} & 0.0626 & 0.0287 & 0.0305  &  \underline{\textbf{0.8934}} & 0.8908 & 0.8480 & 0.8235 & 0.7466 & \underline{0.8912} & 0.8327 & 0.8338 \\
Fetal Heal. &  0.1043 & \underline{0.1662} & 0.0631 & 0.0950 & \underline{\textbf{0.1703}} & 0.1270 & 0.0383 & 0.0510 &  0.9439 & \underline{\textbf{0.9754}} & 0.9339 & 0.9574 & \underline{0.9731} & 0.9486 & 0.9154 & 0.9296  \\
Flare-F  &  0.0161 & 0.2004 & 0.0106 & \underline{\textbf{0.9723}} & \underline{0.3909} & 0.0769 & 0.0029 & \textbf{0.4571} &  0.1719 & 0.9317 & 0.1123 & \underline{1.0000} & 0.9217 & 0.6193 & 0.0298 & \underline{\textbf{1.0000}} \\
glass5  &  \underline{0.3876} & 0.0588 & 0.0518 & 0.0360 & 0.0765 & \underline{\textbf{0.4171}} & 0.0258 & 0.0618 &  \underline{0.9773} & 0.8899 & 0.9007 & 0.8701 & 0.9088 & \underline{\textbf{0.9899}} & 0.8745 & 0.8939 \\
glass6  &  \underline{0.3881} & 0.0602 & 0.0532 & 0.0370 & 0.0800 & \underline{\textbf{0.4167}} & 0.0268 & 0.0750 &  \underline{0.9774} & 0.8959 & 0.8754 & 0.8687 & 0.8877 & \underline{\textbf{0.9909}} & 0.8780 & 0.8896 \\
Heart Dis. &  0.1224 & 0.1103 & 0.0880 & \underline{\textbf{0.4854}} & 0.0963 & \underline{0.1406} & 0.0548 & 0.1017 &  0.8492 & 0.8190 & 0.7869 & \underline{\textbf{0.9901}} & 0.7988 & \underline{0.8667} & 0.7755 & 0.8257  \\
N. Thyroid  &  \underline{0.0823} & 0.0322 & 0.0297 & 0.0226 & 0.0249 & \underline{\textbf{0.1122}} & 0.0175 & 0.0226 &  \underline{0.8754} & 0.8155 & 0.8322 & 0.8057 & 0.8075 & \underline{\textbf{0.9034}} & 0.7999 & 0.8206 \\
yeast  &  0.0320 & \underline{0.2112} & 0.0206 & 0.0219 & \underline{\textbf{0.2636}} & 0.0397 & 0.0197 & 0.0219 &  0.8797 & \underline{0.9184} & 0.8492 & 0.8504 & \underline{\textbf{0.9526}} & 0.9093 & 0.8516 & 0.8631 \\
yeast1  &  0.0347 & \underline{0.2082} & 0.0199 & 0.0191 & \underline{\textbf{0.2622}} & 0.0485 & 0.0158 & 0.0244 &  0.8945 & \underline{0.9141} & 0.8519 & 0.8473 & \underline{\textbf{0.9533}} & 0.9039 & 0.8442 & 0.8644 \\
\hline\hline
\multirow{2}{*}{\textbf{Rank}}  & 3.21 (1-4-0) & 3.29 (3-3-0) & 5.85 (0-0-0) & 5.32 (3-0-1) & \underline{2.93} \underline{(3-3-0)} & \underline{\textbf{2.36}} \underline{\textbf{(4-3-0)}} & 7.93 (0-13) & 5.11 (0-1-0) & \underline{3.29} \underline{(1-5-0)} & 3.50 (2-2-0) & 5.50 (0-0-1) & 5.61 (2-1-3) & 4.50 (2-2-3) & \underline{\textbf{2.21}} \underline{\textbf{(6-3-0)}} & 6.93 (0-0-7) & 4.46 (1-1-0) \\\hline
\end{tabular}
\end{center}
\end{table*}

The relevant literature primarily uses two distance metrics to capture memorization: Distance to 
the Nearest Neighbour (DNN), and Nearest Neighbour Distance Ratio (NNDR) \cite{acml2021,icml2023}. 
DNN is defined as the Gower distance of a synthetic sample from its nearest real neighbour, whereas
NNDR is computed by dividing DNN with the Gower distance of the same synthetic sample from its
second-closest real record. In general, lower DNN and NNDR ratios are indicative of higher 
disclosure risks, while higher ratios imply better privacy.

\begin{figure*}[!t]
\center
\includegraphics[width=.32\linewidth]{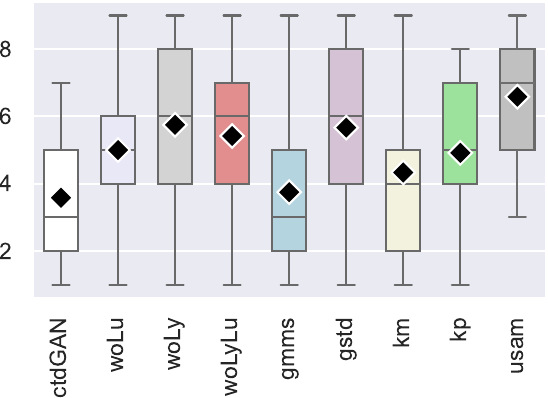}\hspace{7pt}
\includegraphics[width=.32\linewidth]{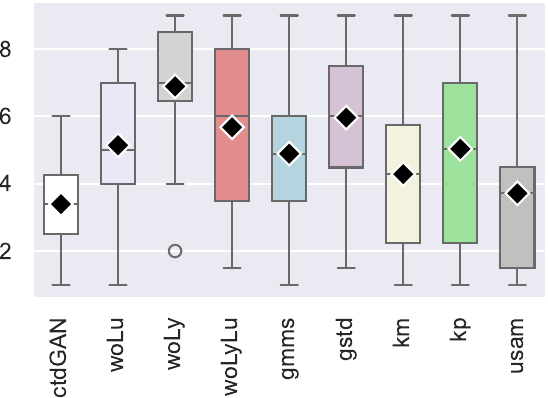}\hspace{7pt}
\includegraphics[width=.32\linewidth]{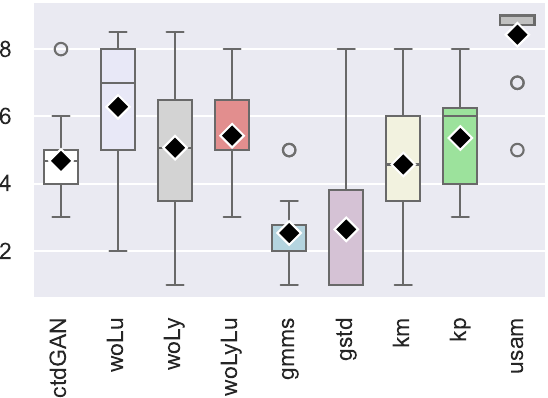}
\caption{
Ablation study results. Left/Middle: Rankings of representative \ganname{} variants in terms of
classification improvement and data fidelity with XGBoost $F1$ scores. Right: Rankings of the same 
variants in terms of privacy preservation/memorization in terms of Nearest Neighbour Distance Ratio
(NNDR).
}
\label{fig:exp-abs}
\end{figure*}

The memorization experiments are challenging because they oppose the data fidelity measurements.
Specifically, high DNN and NNDR values may indicate excellent privacy preservation, but they could 
also reveal unrealistic data generation. Indeed, a dummy model that randomly creates samples would 
achieve top performance here. On the other hand, a hypothetical model that mimics, or even copies
its training set would achieve top data fidelity performance, but it would fail in the DNN and NNDR
experiments.

For this reason, the results of Table~\ref{tab:prv} should be carefully combined with the 
experiments of the previous subsections. For example, SB-GAN and C-GAN scored the highest DNN and
NNDR values here. These measures alone indicate an excellent privacy preservation behavior, but
according to the fidelity experiments, these models were also found to be generating samples that
do not resemble the real ones. On the other hand, TVAE, a model that generated high fidelity data,
was ranked last in both DNN and NNDR tests. This is a signal that the model tends to memorize its
input data and does not generalize well.

In contrast, \ganname{} exhibited a more balanced behavior compared to the evaluated models. 
Specifically, it was proved more robust in terms of privacy preservation than CTAB-GAN+, TVAE, 
ctGAN, and Gaussian Copula. On the other hand, Copula GAN seems to be more private than \ganname{}, 
since its synthetic samples were reasonably far away from their corresponding real ones. However,
as it was shown in the previous subsection, the synthetic samples of Copula GAN were of lower
fidelity than those produced by \ganname{}.

\subsection{Ablation Study}
\label{ssec:exp-abst}

This ablation study investigates the contribution of the main components to the performance of
\ganname{}. In particular, we evaluate eight variants in which specific components of the model are 
removed:

\begin{itemize}
\item \ganname{}-km and \ganname{}-kp replace the hierarchical Agglomerative clustering (HAC)
algorithm with $k$-Means++ and $k$-Prototypes, respectively.

\item \ganname{}-gstd and \ganname{}-gmms replace the cluster-based min-max normalizer of
Eq.~\ref{eq:mms}. The former employs a standard scaler fitted globally on the entire training set,
whereas the latter uses a global min-max scaler.

\item \ganname{}-woLy, \ganname{}-woLu, and \ganname{}-woLyLu remove the loss terms
$\loss_\mathbf{y}$ (Eq.~\ref{eq:gy-loss}), $\loss_\mathbf{u}$ (Eq.~\ref{eq:gu-loss}), and both 
terms, respectively.

\item \ganname{}-usam replaces the probabilistic sampling procedure of
Subsection~\ref{ssec:train-samp} with a uniform cluster sampling strategy that ignores the
probability matrix of Eq.~\ref{eq:pmatrix}.
\end{itemize}

Figure~\ref{fig:exp-abs} summarizes the results across three evaluation dimensions: classification 
performance (left), data fidelity (middle), and privacy preservation (right). The first two
experiments use the  XGBoost $F1$ score, while the third employs NNDR. The boxplots depict the
ranking distributions of the variants, averaged across 5 CV folds, and \numseeds{} random seeds.
The mean and median rankings are respectively indicated by the black markers and the horizontal
lines inside the boxes.

Regarding classification performance, the probabilistic sampling strategy was the most influential
component. Here the \ganname{}-usam variant achieved the lowest ranking among all configurations. 
The proposed loss function also improves performance, as evidenced by the poor ranking of
\ganname{}-woLyLu. In contrast, replacing HAC with $k$-Means resulted in only a marginal 
performance decrease, suggesting that the framework is robust to the choice of clustering 
algorithm.

The fidelity and privacy results should be considered jointly. In terms of fidelity, \ganname{}
achieved the best overall ranking, with the proposed loss formulation and cluster-based
normalization scheme providing the largest contributions. Regarding privacy preservation,
\ganname{}-gmms and \ganname{}-gstd appear to offer the strongest privacy guarantees. However, this
advantage comes at the cost of substantially reduced data fidelity, as these variants were ranked
last in the fidelity evaluation. \ganname{}-usam was by far the weakest variant. This suggests that
the inappropriate cluster selection forces the model to produce samples that closely resemble
training instances.

\begin{figure*}[!t]
\center
\includegraphics[width=.30\linewidth]{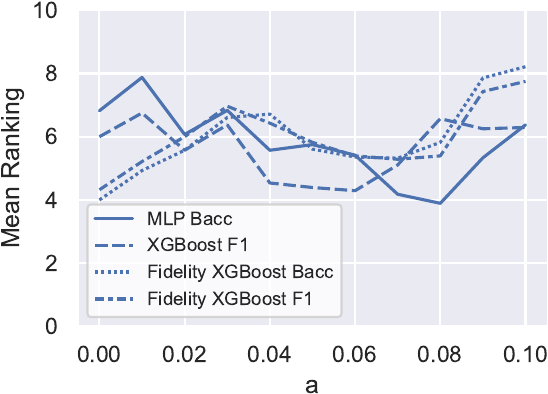}\hspace{7pt}
\includegraphics[width=.67\linewidth]{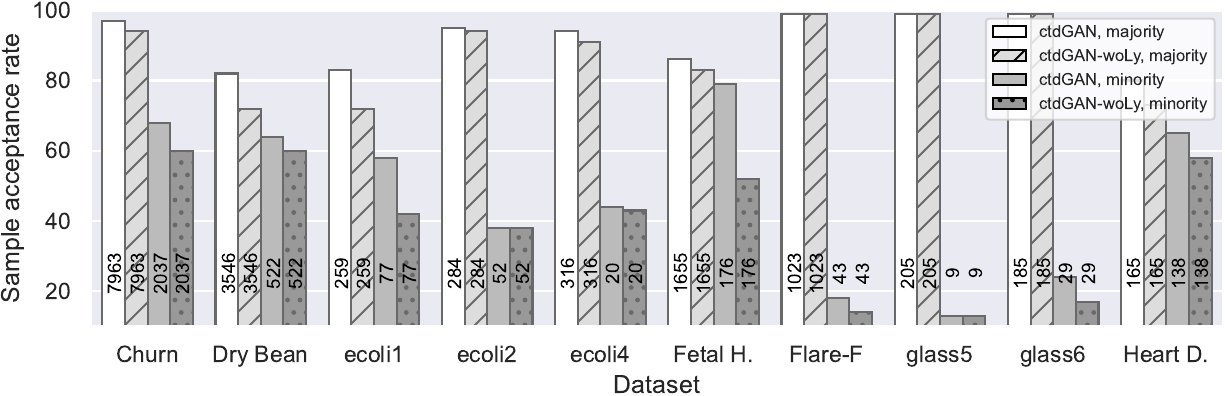}
\caption{
Left: Performance fluctuation of \ganname{} vs $\alpha$ (Eq.~\ref{eq:si}). Right: The impact of
$\loss_\mathbf{y}$ (Eq.~\ref{eq:gy-loss}) in the conditional sample generation ability of 
\ganname{}.
}
\label{fig:exp-abs2}
\end{figure*}

\begin{table}[!t]
\begin{center}
\caption{Selected number of clusters and cluster distribution for the classification experiment 
(median, min and max values).}
\label{tab:hpa}
\begin{tabular}{| l | c | P{1.5cm} | P{1.5cm} | P{1.5cm} |}            \hline
\multirow{2}{*}{\bf Dataset} & \multirow{2}{*}{$k$} & {\bf Smallest cluster (\%)} & {\bf Median cluster (\%)} & {\bf Largest cluster (\%)} \\\hline
 Anemia & 2 (2-2) & 9.07 & 9.07 & 91.93 \\
 Churn & 5 (5-7) & 7.59 & 19.44 & 27.36 \\
 DryBean & 5 (5-5) & 2.01 & 16.47 & 46.98 \\
 ecoli1 & 4 (4-5) & 2.60 & 27.05 & 43.12 \\
 ecoli2 & 4 (4-5) & 2.97 & 26.02 & 44.24 \\
 ecoli4 & 4 (4-6) & 2.60 & 26.77 & 41.64 \\
 Fetal H. & 4 (4-5) & 4.06 & 24.93 & 44.33 \\
 flare & 3 (3-4) & 1.88 & 22.54 & 70.34 \\
 glass5 & 4 (3-4) & 3.51 & 10.47 & 76.61 \\
 glass6 & 4 (3-4) & 2.92 & 12.28 & 73.84 \\
 Heart & 5 (3-5) & 6.58 & 21.07 & 37.04 \\
 N. Thyr. & 5 (4-6) & 2.33 & 7.85 & 78.49 \\
 yeast & 7 (6-8) & 0.25 & 6.15 & 50.67 \\
 yeast1 & 6 (4-7) & 0.84 & 5.13 & 61.08 \\\hline
\end{tabular}
\end{center}
\end{table}

In the sequel, we examine the impact of the proposed loss function ($\loss_\mathbf{y}$ term), in
the ability of \ganname{} to conditionally generate samples. The right part of Fig.
\ref{fig:exp-abs2} compares the sample acceptance rates of \ganname{} vs the \ganname{}-woLy
variant. Recall that a sample is acceptable only if its generated class agrees with the requested
class label (step 13 of Algorithm \ref{algo:ps}). These measurements were obtained during the
fidelity test, where the models attempted to construct a synthetic dataset that resembles the
original one. The numbers inside the bars indicate the number of samples that were generated from
the majority and minority classes. The diagram shows that the auxiliary classifier $\clfqy$ along with the loss term $\loss_\mathbf{y}$ lead to substantial improvements in conditional sample
generation.

Furthermore, the left diagram of Fig.~\ref{fig:exp-abs2} plots the performance variations of our
model vs. fluctuations of the $\alpha$ hyper-parameter of Eq.~\ref{eq:si}. The figure reveals that in general, \ganname{} achieves a quite stable performance in the range $\alpha\in[0.03,0.08]$.
According to the diagram, \ganname{} achieved high performances in all classification tests with
$\alpha=0.07$; therefore, we chose this value in this study.

We also studied how $\alpha$ affects the clustering process of \ganname{}. Table~\ref{tab:hpa} 
presents the cluster statistics across the cross-validation runs of the classification improvement 
experiment. The selected number of clusters, $k$, is reported as median (minimum-maximum) across 
all training folds and random states. For each training subset, the minimum, median, and maximum 
relative cluster sizes were computed, and the table reports the median of each statistic across all 
experiments.

Finally, Table~\ref{tab:exp-ss} contains the results of some additional statistical significance
tests for several cases. The presented $p$-values reflect the results of the non-parametric Finner
test. They were all found to be statistically significant, assuming a significance level equal to
$5\cdot 10^{-2}$.

\begin{table}[!t]
\begin{center}
\caption{Model training times in seconds. The parenthesized numbers in \ganname{} denote the clustering duration.}
\label{tab:times}
\setlength\tabcolsep{3pt}
\begin{tabular}{| l | P{0.85cm} | P{0.7cm} | P{0.9cm} | P{0.7cm} | P{0.7cm} | P{0.75cm} | P{1.4cm} |} \hline
\multirow{2}{*}{\bf Dataset}  & \multirow{2}{*}{\bf CGAN} & {\bf COP GAN} & {\bf CTAB-GAN+} & {\bf CT GAN} & {\bf SB GAN} & \multirow{2}{*}{\bf TVAE} & \multirow{2}{*}{\bf \ganname{}} \\\hline
Anemia & 7.2 & 31.4 & 130.1 & 30.9 & 6.4 & 20.4 & 21.9 (2.8) \\
Churn & 51.1 & 191.1 & 993.3 & 191.0 & 48.3 & 77.8 & 190.9 (19.0) \\
DryBean & 69.0 & 328.3 & 1698.4 & 331.4 & 66.2 & 184.3 & 235.8 (17.2)\\
ecoli1 & 2.1 & 4.9 & 14.2 & 4.7 & 2.1 & 3.4 & 3.6 ($<$0.1) \\
ecoli2 & 2.1 & 4.9 & 14.2 & 4.7 &  2.1 & 3.4 & 3.7 ($<$0.1) \\
ecoli4 & 2.3 & 5.2 & 16.2 & 4.9 & 2.2 & 4.0 & 3.9 ($<$0.1) \\
Fetal H. & 11.2 & 62.9 & 258.6 & 64.0 & 11.1 & 37.1 & 42.9 (1.6) \\
flare & 6.3 & 20.8 & 105.1 & 20.5 &  6.2 & 6.6 & 25.3 ($<$0.1) \\
glass5 & 1.4 & 3.1 & 11.3 & 2.8 &  1.4 & 13.8 & 2.1 ($<$0.1) \\
glass6 & 1.4 & 3.0 & 10.9 & 2.8 & 1.4 & 2.8 & 2.0 ($<$0.1)\\
Heart & 2.1 & 5.9 & 24.8 & 5.8 &  2.1 & 3.7 & 5.0 ($<$0.1)\\
N. Thyr. & 1.4 & 2.6 & 6.7 & 2.4 & 1.4 & 2.1 & 1.9 ($<$0.1)\\
yeast & 7.9 & 27.2 & 139.3 & 26.3 & 6.6 & 29.7 & 22.2 (0.1) \\
yeast1 & 8.2 & 26.2 & 135.5 & 25.7 & 8.1 & 15.9 & 20.2 (0.1) \\\hline
\end{tabular}
\end{center}
\end{table}

\subsection{Training Times}
\label{ssec:exp-tt}

Table~\ref{tab:times} illustrates the model training times. The parenthesized numbers next to
\ganname{} fit times indicate the duration of the clustering step. This time is part of the
training phase time (i.e. it is not added to it) and ranges from 2\% to 13\%. The results indicate
the impact of clustering in the total training process of the \ganname{} ranges from infinitesimal
to modest. Overall, \ganname{} is faster than many state-of-the-art networks like ctGAN, CTAB-GAN+,
and Copula GAN, but it is slower compared to simple baselines like SB-GAN and the vanilla 
conditional GAN.

\begin{table*}[!t]
\begin{center}
\caption{Indicative non-parametric statistical significance results of \ganname{} vs. other models 
for various cases with the Finner test}
\label{tab:exp-ss}
\setlength\tabcolsep{5.5pt}

\begin{tabular}{| l | c | c | c | c | c | c | c | c | c | c |}               \hline
{\bf Model} & {\bf CGAN} & {\bf COPGAN} & {\bf CTAB-GAN} & {\bf CTGAN} & {\bf GCOP} & {\bf SB-GAN} & {\bf TVAE} \\\hline
Classification, XGBoost $F1$ & $2.77\cdot 10^{-2}$ & $7.65\cdot 10^{-3}$  & $3.87\cdot 10^{-3}$  & $9.57\cdot 10^{-3}$  & $5.20\cdot 10^{-4}$  & $2.08\cdot 10^{-2}$  & $9.13\cdot 10^{-3}$  \\
Classification, XGBoost $B_{ac}$ & $3.83\cdot 10^{-3}$ & $2.01\cdot 10^{-4}$  & $2.97\cdot 10^{-5}$  & $3.13\cdot 10^{-3}$  & $2.97\cdot 10^{-5}$  & $5.72\cdot 10^{-3}$  &  $9.25\cdot 10^{-3}$  \\
Fidelity, XGBoost $B_{ac}$  & $4.34 \cdot 10^{-8}$  &  $1.05\cdot 10^{-2}$  &  $4.21\cdot 10^{-3}$  & $8.20\cdot 10^{-3}$  & $2.19\cdot 10^{-5}$  & $2.66\cdot 10^{-8}$ & $2.95\cdot 10^{-2}$  \\\hline
\end{tabular}
\end{center}
\end{table*}

\section{Discussion and Limitations}
\label{sec:disc}

Our experimental evaluation on a diverse collection of \numdatasets{} datasets demonstrated the
usefulness of \ganname{} across a wide range of scenarios. Although several competing methods
occasionally achieved better performance on individual datasets, our proposed model was on average
more effective in terms of classification performance and generated data fidelity.

Notably, the inclusion of \numdatasets{} datasets makes this study one of the most extensive
evaluations in the relevant literature. In addition, the robustness of the experimental protocol --
5-fold cross-validation, repeated executions using \numseeds{} random seeds, and statistical
significance testing -- enhances the reliability and credibility of the reported results.

Tables~\ref{tab:os}--\ref{tab:ccorr} demonstrate that synthesizing samples within carefully
selected subspaces of the original feature space leads to substantial improvements in both 
classification performance and data fidelity. Remarkably, \ganname{} was the only deep learning
model that outperformed both SMOTE and ADASYN across all classification experiments. On average, it
achieved the highest ranking by a considerable margin.

TVAE was the strongest opponent in the fidelity experiment, indicating that Variational 
Autoencoders may have a potential in tabular data generation tasks. ctGAN and CTAB-GAN+ were also 
competitive in several cases, particularly when they were combined with XGBoost and Random Forest 
classifiers.

In terms of privacy preservation, \ganname{} was less private than SB-GAN, the Conditional GAN and
Copula GAN. Regarding the first two, this is largely explained by the weakness of these models to
generate realistic data. Similarly, the quality of the synthetic data of Copula GAN was also
inferior.

Despite its strong performance, \ganname{} has several limitations. First, it requires an initial
clustering step, which imposes an additional computational cost during training. Moreover, the
model's effectiveness is affected by the selected clustering algorithm. Experiments with $k$-Means
produced results comparable to those obtained with Agglomerative Clustering, whereas the use of 
$k$-Prototypes resulted in a noticeable degradation in performance.

Finally, although the cluster-aware min--max normalization technique presented in 
Subsec.~\ref{ssec:trans} combines simplicity with effectiveness, it remains sensitive to the 
presence of outliers. Incorporating an outlier detection mechanism, similar to that 
employed by SB-GAN, could further improve performance.

\section{Conclusions and Future Work}
\label{sec:conc}

In this paper, we presented \ganname{}, a conditional Generative Adversarial Network for tabular
data synthesis in imbalanced datasets. The main idea is to conditionally synthesize samples that 
not only belong to a particular class, but are also located in locations of the real data space
where it is more possible to encounter samples of that class.

\begin{table}[!b	]
\begin{center}
\caption{Method at a glance.}
\label{tab:mag}

\begin{tabular}{| l | l |} \hline
{\bf Component}  & {\bf Usage/Usefulness} \\\hline
Clustering  & \makecell[l]{Conditionally generate samples not just from a\\particular class, but also in specific regions of the\\feature space where samples of that class are more\\likely to occur.} \\\hline
\makecell[l]{Probabilistic\\sampling} & \makecell[l]{During sampling, a conditional probability matrix is\\used to determine the most suitable clusters for the\\requested class.} \\\hline
\makecell[l]{Loss terms\\$\loss_\mathbf{y}$ and $\loss_\mathbf{u}$} & \makecell[l]{Penalize the generation of samples with unrealistic\\class and cluster labels during training.\\They improve conditional sampling.} \\\hline
\makecell[l]{Cluster-based\\scaling} & \makecell[l]{Captures different modes in the continuous variable\\distributions of the samples of the same cluster without\\increasing dimensionality.}\\\hline
\end{tabular}
\end{center}
\end{table}

To achieve this objective, we introduced two key elements. First, we employ a clustering algorithm
to assign cluster labels to all training samples. These cluster labels are used alongside class
labels during the training process, enabling \ganname{} to generate samples that belong 
simultaneously to a specific class and cluster. Second, we introduce a probabilistic sampling
mechanism. Given the requirement to generate samples for a target class $y$, a probability matrix 
is constructed to identify the clusters that are most suitable for that class. Consequently, the 
single condition \textit{``generate $N$ samples from class $y$''} is transformed into the dual 
condition \textit{``generate $N$ samples from class $y$ and a set of appropriate clusters''}. 
Experimental results on \numdatasets{} benchmark datasets demonstrated that \ganname{} achieved the 
strongest overall classification performance among the evaluated state-of-the-art generative models 
and oversampling techniques.

As future work, we plan to investigate alternative clustering techniques within the probabilistic
sampling framework. In particular, we aim to explore methods for identifying clusters that overlap
with class decision boundaries, as these regions are likely to contain more informative samples.

\bibliographystyle{IEEEtran}
\bibliography{tkde.bib}

%\newpage

\begin{IEEEbiography}[{\includegraphics[width=1in,height=1.25in,clip,keepaspectratio]
{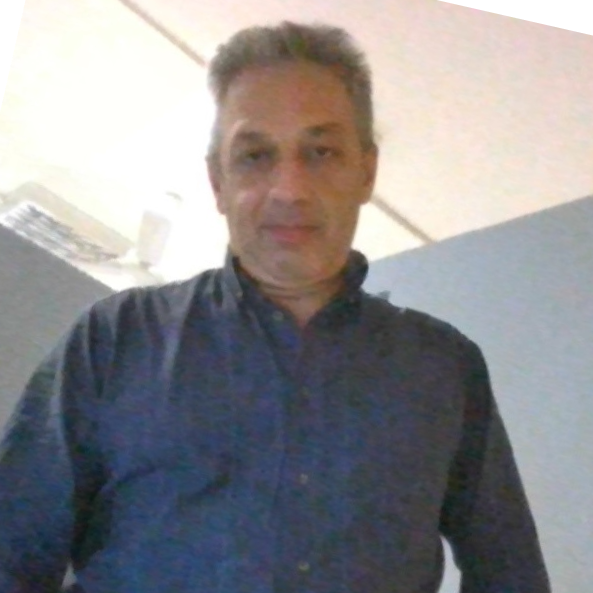}}]{Leonidas Akritidis} is an Assistant Professor in the Department of Information 
and Electronic Engineering of the International Hellenic University. He is also a contracted
lecturer in the Science and Technology Dept. of the same university. He holds a diploma (2003) and
a PhD (2013) in Electrical and Computer Engineering. His research activity is focused on the fields
of Deep Generative Modeling, Natural Language Processing, and Data Engineering. He has published
research articles in leading international journals and scientific conferences. Moreover, he has
designed and developed a broad collection of scientific and commercial systems. He has contributed
to the successful preparation and completion of various research projects with national and
international funding.
\end{IEEEbiography}

\begin{IEEEbiography}[{\includegraphics[width=1in,height=1.25in,clip,keepaspectratio]
{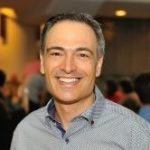}}]{Panayiotis Bozanis} is currently a full Professor at the Science and Technology
Dept., International Hellenic University, Greece, since September, 2019. He holds a Diploma and a
PhD degree in Computer Engineering and Informatics, both from University of Patras, Greece.
Previously, he served as full Professor, Head of the Department, Deputy Dean, Director of the MSc
Programme “Applied Informatics”, and Director of the DaSELab at the ECE Dept., University of
Thessaly, Greece. His publications comprise several journal/conference papers, book chapters, eight
books (in Greek) about Data Structures, Algorithms and Introduction to Computer Science and editing
of seven books. His research interests include, among others, Data Structures, Algorithms,
Information Retrieval, Databases, Cloud Computing, Big Data, Machine Learning, and Smart Grids.
\end{IEEEbiography}

\end{document}